%% file: acl_latex.tex
\pdfoutput=1
\documentclass[11pt]{article}

\usepackage[preprint]{acl}

\usepackage{times}
\usepackage{latexsym}

\usepackage[T1]{fontenc}
\usepackage[utf8]{inputenc}

\usepackage{microtype}

\usepackage{inconsolata}

\usepackage{graphicx}

\usepackage{booktabs}
\usepackage{amsmath,amsfonts}

\usepackage{array}
\usepackage{textcomp}
\usepackage{stfloats}
\usepackage{url}
\usepackage{verbatim}
\usepackage{graphicx}
\usepackage[utf8]{inputenc}
\usepackage{multirow}
\usepackage{multicol}
\usepackage{color}
\usepackage{url}
\usepackage{enumitem}
\usepackage{diagbox}
\usepackage{flushend,cuted}
\usepackage{bbm}
\usepackage{xspace}
\usepackage{algpseudocode}
\usepackage{colortbl}
\usepackage{bbding}
\usepackage{amssymb}

\usepackage{listings}
\usepackage{ulem}
\usepackage{listings}
\usepackage{wrapfig}
\usepackage{soul}
\usepackage{makecell}
\usepackage{xcolor}

\newcommand{\kaiti}[1]{\begin{CJK*}{UTF8}{gkai} #1 \end{CJK*}}
\soulregister{\kaiti}7
\usepackage{pifont}
\newcommand{\mycircled}[1]{%
   \raisebox{2pt}{\textcircled{\raisebox{-0.9pt}{\kern-0.2pt #1}}}%
}
\usepackage{rotating}
\usepackage{stfloats}
\usepackage{CJKutf8}
\usepackage[encapsulated]{CJK}
\usepackage{tabularx}

\newcommand{\name}{BenchMAX\xspace}

\renewcommand{\arraystretch}{1.1}

\title{\includegraphics[width=0.6cm]{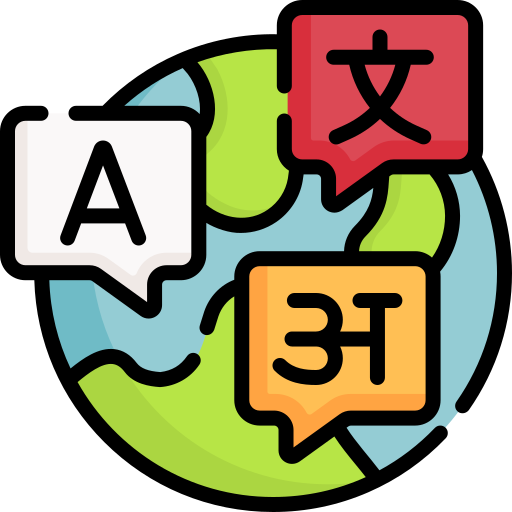}\ \name: A Comprehensive Multilingual Evaluation Suite \\ for Large Language Models}

\author{
 \textbf{Xu Huang\textsuperscript{1}},
 \textbf{Wenhao Zhu\textsuperscript{1}},
 \textbf{Hanxu Hu\textsuperscript{2}},
 \textbf{Conghui He\textsuperscript{3}},
 \textbf{Lei Li\textsuperscript{4}}, \\
 \textbf{Shujian Huang\textsuperscript{1}\thanks{Corresponding authors}},
 \textbf{Fei Yuan\textsuperscript{3}\footnotemark[1]}
\\
 \textsuperscript{1}National Key Laboratory for Novel Software Technology, Nanjing University \\
 \textsuperscript{2}University of Zurich,\textsuperscript{4}Carnegie Mellon University \\
 \textsuperscript{3}Shanghai Artificial Intelligence Laboratory \\
\small{\texttt{\{xuhuang,zhuwh\}@smail.nju.edu.cn, hanxu.hu@uzh.ch, heconghui@pjlab.org.cn}}
\\
\small{\texttt{leili@cs.cmu.edu, huangsj@nju.edu.cn, yuanfei@pjlab.org.cn}}
}

\begin{document}
\maketitle

\input{sections/00_abstract}

\input{sections/01_introduction}

\input{sections/02_related_work}

\input{sections/03_systematic_construction}

\input{sections/04_experiments}

\input{sections/05_analysis}

\input{sections/06_conclusion}

\normalem
\bibliography{custom}

\input{sections/07_appendix}

\end{document}

%% file: sections/00_abstract.tex
\begin{abstract}
Previous multilingual benchmarks focus primarily on simple understanding tasks, but for large language models~(LLMs), we emphasize proficiency in instruction following, reasoning, long context understanding, code generation, and so on.
However, measuring these advanced capabilities across languages is underexplored.
To address the disparity, we introduce \name, a multi-way multilingual evaluation benchmark that allows for fair comparisons of these important abilities across languages. 
To maintain high quality, three distinct native-speaking annotators independently annotate each sample within all tasks after the data was machine-translated from English into 16 other languages. 
Additionally, we present a novel translation challenge stemming from dataset construction.
Extensive experiments on \name reveal varying effectiveness of core capabilities across languages, highlighting performance gaps that cannot be bridged by simply scaling up model size.
\name serves as a comprehensive multilingual evaluation platform, providing a promising test bed to promote the development of multilingual language models. The dataset\footnote{\url{https://huggingface.co/collections/LLaMAX/benchmax-674d7a815a57baf97b5539f4}} and code\footnote{\url{https://github.com/CONE-MT/BenchMAX.git}} are publicly accessible.

% mere model scaling fails to resolve. 
% \name pays more attention to multilingual variations of LLM core capabilities, which provides a promising test bed and hence promotes further development to bridge the gap between languages~\footnote{The dataset and evaluation code will be publicly accessible}.

% Constructing appropriate benchmarks to evaluate the key capabilities of large language models is crucial for further development. There is a series of works targeting the core capabilities of large language models such as math and code reasoning, long context understanding, and instruction following, 
% We find that even among the advanced large language models, there remains a significant performance gap across different languages, and merely scaling the model size does not consistently mitigate the gap. Our work provides a promising multilingual test bed for evaluating frontier capabilities and hence promotes further development to bridge the gap between languages.
\end{abstract}

%% file: sections/01_introduction.tex
\section{Introduction}

Large Language Models~(LLMs; ~\citealp{openai2024gpt4,team2024gemini,DeepSeekAI2024DeepSeekV3TR}) have displayed remarkable proficiency across a wide range of tasks, mainly because they excel in instruction following, reasoning, long context understanding, code generation, and so on~\cite{ouyang2022training,cobbe2021training,su2024roformer,roziere2023code,lu2024llamax,sun2024survey}. Inherently, these capabilities are language-agnostic. 
Consider a simple task like the acquisition of mathematical concept: the numerical outcome remains consistent regardless of whether one learns the arithmetic expression $1+1=2$ in English or Chinese.  
% Taking the acquisition of mathematical concepts as an example, whether learning arithmetic expressions in English or Chinese, the numerical meaning is the same.
Similarly, when it comes to coding tasks, the choice between English or Chinese for articulating these instructions does not alter the fundamental logic of the code. 
However, numerous empirical studies have shown that LLMs' multilingual performance is quite unbalanced across different languages when handling same tasks~\cite{shi2023language, zhu2024multilingual, qi2023cross}.

\begin{figure}[t]
    \centering
    \includegraphics[width=0.9\linewidth]{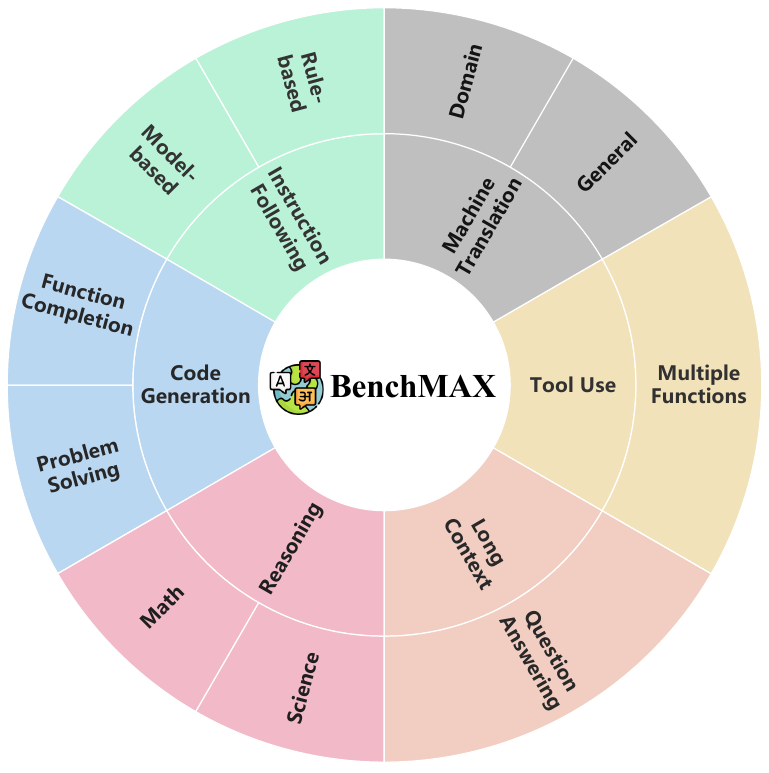}
    \caption{\name evaluates diverse advanced capabilities of LLMs in multilingual context.}
    \label{fig:capability}
    \vskip -0.2in
\end{figure}

However, current benchmarks~\cite{hendrycks2021measuring,lai2023okapi,singh2024global,wang2024seaeval} do not support comprehensive testing of the language-agnostic abilities of LLMs, particularly in low-resource language settings, for several reasons. Tasks like XWinograd~\cite{xwinograd-muennighoff-etal-2023-crosslingual} and XStoryCloze~\cite{xstorycloze-lin-etal-2022-shot}, based on multiple-choice formats, do not fully evaluate the generative capacities of LLMs. 
Additionally, the limited language overlap across existing benchmarks poses challenges in assessing LLM performance in diverse languages. 
Recently, P-MMEval~\cite{zhang2024pmmeval} is proposed as a multilingual multitask benchmark, with the majority of its tasks still following a multiple-choice format. 
While it includes assessments like MGSM~\cite{shi2023language} and MIFEVAL that cover partial language-agnostic capabilities, LLMs exhibit remarkable performance, as shown in Table~\ref{tab:benchmark_comp}. 
This narrow focus leaves a significant gap between research evaluation and real-world applications.

To tackle this problem, we develop a comprehensive, multi-way, and challenging multilingual evaluation suite, called \name, to help the community better analyze and improve the language-agnostic capabilities of LLMs. 
Covering 17 languages\footnote{The 17 languages include English, Spanish, French, German, Russian, Bengali, Japanese, Thai, Swahili, Chinese, Telugu, Arabic, Korean, Serbian, Czech, Hungarian, and Vietnamese.}, \name not only includes a broader range of language families but also emphasizes the diversity of writing systems across languages. 
As demonstrated in Table~\ref{tab:benchmark_comp}, \name increases the percentage of studied languages that utilize the non-Latin script.

\begin{table}[t]
    \centering
    \setlength{\tabcolsep}{2pt}
    \scriptsize
    \resizebox{\linewidth}{!}{
    \begin{tabular}{l|cc|ccc}
    \toprule
         \textbf{Tasks} & \makecell{\textbf{Llama3.1}\\ \textbf{70B}} & \makecell{\textbf{Qwen2.5}\\ \textbf{72B}} & \makecell{\textbf{\# LG} \\ } & \makecell{\textbf{\# LG-Family} \\ \textbf{Diversity}} & \makecell{$\boldsymbol{R_{\mathrm{non-Latin}}}$}\\
    \midrule
        XWinograd & 69.7 & 83.7 & 6 & 3 & 50.0 \\
        XStoryCloze & 70.3 & 83.6 & 13 & 11 & 38.5 \\
        MGSM$^*$  & 88.3 & 79.2 & 10 & 7 & 50.0 \\
        MIFEVAL$^*$ & 91.0 & 87.6 & 10 &  7 & 50.0 \\
        \midrule
        \multicolumn{6}{l}{\name} \\
        \midrule
        - Instruction Following & 11.1 & 34.1 & \multirow{4}{*}{17} & \multirow{4}{*}{11} & \multirow{4}{*}{58.8} \\
        - Code Generation & 29.8 & 45.5 \\
        - Science Reasoning & 35.8 & 39.4 \\
        - Tool Use & 44.3 & 61.8 \\
    \bottomrule
    \end{tabular}
    }
    \caption{\name provides a more comprehensive analysis of LLM language-agnostic capabilities by covering a broader range of capability scenarios, language families, and script systems. \# LG and \# LG-Family denote the number of supported languages and the language families they belong to, respectively. $R_{\mathrm{non-Latin}}$ refers to the proportion of languages that do not use the Latin script among supported languages. $*$ The results are from P-MMEval. }
    \label{tab:benchmark_comp}
    \vskip -0.2in
\end{table}

Meanwhile, \name highlights diverse language-agnostic advanced capabilities  (Figure~\ref{fig:capability}). 
We assess instruction following capability with rule-based~\cite{zhou2023instruction} and model-based~\cite{li2024crowdsourced} evaluations, code generation capability in diverse scenarios~(function-completion~\cite{liu2024your}~/~problem solving~\cite{jain2024livecodebench}), long context understanding capability~\cite{hsieh2024ruler}, a verity of reasoning in math~\cite{shi2023language} and science~\cite{Rein2023GPQAAG}, tool use~\cite{srinivasan2023nexusraven} in agent environments, and general~\cite{costa2022no}~/~domain translation. 
Domain translation, a by-product of data construction, poses a new challenge for LLM by necessitating fine-grained control and domain-specific terminology understanding over the translation process.

To ensure high quality, we devise an annotation framework to optimize the dataset quality with human effort and LLM feedback. 
The process involves translating data from English to selected non-English languages using machine translation systems, post-editing each sample by three native-speaking annotators with multiple iterations across most tasks, and picking the final translation version using a strong LLM that involves swapping sample positions for debiasing~\cite{wang2024large,li2024crowdsourced}.

Leading multilingual LLMs are evaluated on \name, revealing that language significantly influences language-agnostic capabilities of existing LLMs. 
Interestingly, simply increasing the parameters can boost average performance on these tasks but does not universally reduce the performance gap across languages. 
Moreover, compared to general translation, domain translation not only poses new challenges for LLMs but also requires new evaluation metrics. 
The main contributions can be summarized as follows:
\begin{itemize} [nosep,itemsep=1pt,leftmargin=0.3cm]
    \item We develop a comprehensive, multi-way multilingual benchmark across \textbf{17} languages for evaluating \textbf{6} crucial capabilities on \textbf{10} diverse tasks.
    \item We propose a pipeline for curating high-quality mutlilingual datasets, involving both human annotation and LLM-as-a-judge.
    \item We evaluate leading multilingual LLMs on \name, and the related analyses provide a further understanding of the language-agnostic capabilities.
\end{itemize}

%% file: sections/02_related_work.tex
\begin{table*}[!ht]
    \centering
    \scriptsize
    \resizebox{\linewidth}{!}{
    \begin{tabular}{c|c|c|c|c|c|c|c}
    \toprule
         \textbf{Language} & \textbf{ISO} & \textbf{Language Family} & \textbf{Script System} & \textbf{Language} & \textbf{ISO} & \textbf{Language Family} & \textbf{Script System} \\
        \midrule
        Hungarian & hu & Uralic & \multirow{6}{*}[-2ex]{Latin} & Serbian & sr & Indo-European & Serbian Cyrillic  \\
        \cline{1-3} \cline{5-8}
        Vietnamese & vi & Austroasiatic & ~ & Korean & ko & Koreanic & Hangul / Chosŏn'gŭl \\
        \cline{1-3} \cline{5-8}
        Spanish & es & \multirow{6}{*}{Indo-European} & ~  & Japanese & ja & Japonic & \makecell{ Mixed scripts of \\ Chinese Characters \\and Hiragana, Katakana} \\
        \cline{1-2} \cline{5-6} \cline{7-8}
        Czech & cs & ~ & ~ & Arabic & ar & Afro-Asiatic & Arabic alphabet \\
        \cline{1-2} \cline{5-8}
        French & fr & ~ & ~ &Thai & th & Kra–Dai & Thai \\
        \cline{1-2} \cline{5-6}  \cline{7-8}
        German & de & ~ & ~ & Swahili & sw & Niger–Congo & Latin \\
        \cline{1-2} \cline{5-6}  \cline{4-4} \cline{7-8}
        Russian & ru & ~ & Cyrillic &  Chinese & zh & Sino-Tibetan & Chinese Characters    \\
        \cline{1-2} \cline{5-6}  \cline{4-4} \cline{7-8}
        Bengali & bn & ~ & Bengali–Assamese&Telugu & te & Dravidian & Telugu \\
    \bottomrule
    \end{tabular}}
    \caption{Besides English, \name supports 16 non-English languages, covering a wide range of language families and script systems.}
    \label{tab:lg_selection}
    % \vskip -0.1in
\end{table*}

\begin{table*}[ht]
    \centering
    \scriptsize
    \setlength{\tabcolsep}{3pt}
    \resizebox{\linewidth}{!}{
    \begin{tabular}{c|c|c|c|c||c|c|c|c|c}
    \toprule
        \textbf{Capability} & \textbf{Category} & \textbf{Dataset} & \textbf{\# Samples} & \textbf{Metric} & \textbf{Capability} & \textbf{Category} & \textbf{Dataset} & \textbf{\# Samples} & \textbf{Metric} \\ 
    \midrule
        Instruction & Rule-based & IFEval & 429 & Accuracy  & \multirow{2}[3]{*}{Code Generation} & \makecell{Function\\Completion} & Humaneval+ & 164 & \multirow{2}[3]{*}{Pass@1}  \\ 
        \cline{2-5} \cline{7-9}
        Following & Model-based & m-ArenaHard & 500 & Win Rate &  & \makecell{Problem\\Solving} & LiveCodeBench\_v4 & 713 & \\
        \hline
        \multirow{2}{*}{Reasoning} & Math & MGSM & 250 & \multirow{2}{*}{Exact Match} & \multirow{2}{*}{Translation} & General & Flores+TED+WMT24 & $[$1012, 4049$]$ & \multirow{2}{*}{spBLEU} \\
        \cline{2-4} \cline{7-9}
        & Science & GPQA & 448 &  & & Domain & Annotated data above& 2781 &  \\
        \hline
        Tool Use & \makecell{Multiple\\Functions} & Nexus & 318 & Accuracy &  Long Context Modeling & \makecell{Question\\Answering} & RULER & 800 & Exact Match \\
    \bottomrule
    \end{tabular}}
    \caption{Selection of core capabilities and details of task data. For IFEval, we filter out all language-specific instructions, thus remaining 429 samples. For Nexus, we only adopt the standardized\_queries subset which contains 318 samples. For general translation datasets, the number of samples may vary in different translation directions, according to the number of parallel samples in TED and WMT24. The datasets of the model-based instruction following task and math reasoning are expanded from existing multilingual datasets, while others are translated from English datasets.}
    \label{tab:task_detail}
    \vskip -0.1in
\end{table*}

\section{Related Work}
% \subsection{Expanding LLM's Multilingual Support} % in Large Language Models
% Various training strategies have been proposed to empower LLM on multilingual scenarios, raning from pre-training~\cite{he2024scaling,alves2024tower} to post-training~\cite{zhu2024power,she2024mapo}.
% In contrast to existing efforts focused on constructing multilingual training datasets~\cite{singh2024aya, li2023bactrian, lai2023okapi}, 
Prior to the era of Large Language Models~~(LLMs; ~\citealp{openai2024gpt4,team2024gemini,DeepSeekAI2024DeepSeekV3TR}), most multilingual benchmarks are designed to evaluate discriminative models and take the form of classification tasks, such as \textsc{XNLI}~\cite{conneau2018xnli}, \textsc{XCOPA}~\cite{ponti2020xcopa}, \textsc{XCSQA}~\cite{talmor2019commonsenseqa}, and so on~\cite{xstorycloze-lin-etal-2022-shot,xwinograd-muennighoff-etal-2023-crosslingual}.
However, due to their limited task complexity and the lack of diversity in format, these tasks become less practical.
% In this paper, we focus on curating multilingual benchmarks tailored for LLM evaluation.
% \noindent\paragraph{Benchmarking general capability}
Recently, MGSM has become the most frequently used dataset in papers and reports from leading LLM teams~\cite{dubey2024llama,team2024gemini,openai2024gpt4o}, which measures the mathematical reasoning capability across eleven languages.
In this paper, we extend it to cover six additional diverse languages. % , further broadening its scope.
Another widely used benchmark is the multilingual translated version of MMLU~\cite{hendrycks2021measuring,lai2023okapi,singh2024global}, which assesses LLMs on knowledge-intensive tasks. 
However, due to the lack of a unified dataset version, scores are often difficult to compare across studies. 
Moreover, recent analyses have revealed that MMLU contains numerous ground truth errors~\cite{gema2024are}, obscuring the accurate evaluation.
More recently, \textsc{Include}~\cite{romanou2024include} is proposed to evaluate the multilingual regional knowledge, hence it's questions are not language-agnostic.
To address these limitations, our work builds upon GPQA dataset~\cite{Rein2023GPQAAG} instead of MMLU, which offers higher-quality annotations and poses greater challenges in domain-specific knowledge and reasoning evaluation. 
In addition to curating multilingual versions of MGSM and GPQA, we incorporate a broader range of capabilities, including long context modeling~\cite{hsieh2024ruler}, tool use~\cite{srinivasan2023nexusraven}, instruction following~\cite{zhou2023instruction}, and more. 
This allows our benchmark to evaluate LLMs' multilingual capabilities more comprehensively compared to previous aggregated benchmarks, such as SeaEval~\cite{wang2024seaeval} and P-MMEval~\cite{zhang2024pmmeval}.
More importantly, all translations except the long context data in our benchmark are post-edited by native human experts after machine translation.
This significantly improves both the quality and reliability of the dataset.

%% file: sections/03_systematic_construction.tex
\section{Benchmark Construction}
In this section, we extend the evaluation of the core capabilities of LLMs into multilingual scenarios.
To ensure sufficient linguistic diversity, we select 16 non-English languages~(\S~\ref{sec:lg_selection}).
Meanwhile, a diverse set of tasks designed to evaluate 6 crucial LLM capabilities is chosen to facilitate comprehensive assessment~(\S~\ref{sec:capability_selection}).
Subsequently, we introduce a rigorous pipeline~(\S~\ref{sec:construction_process}) that incorporates human annotators and LLMs to obtain a high-quality benchmark.

\begin{figure*}[ht]
    \centering
    \includegraphics[width=0.8\linewidth]{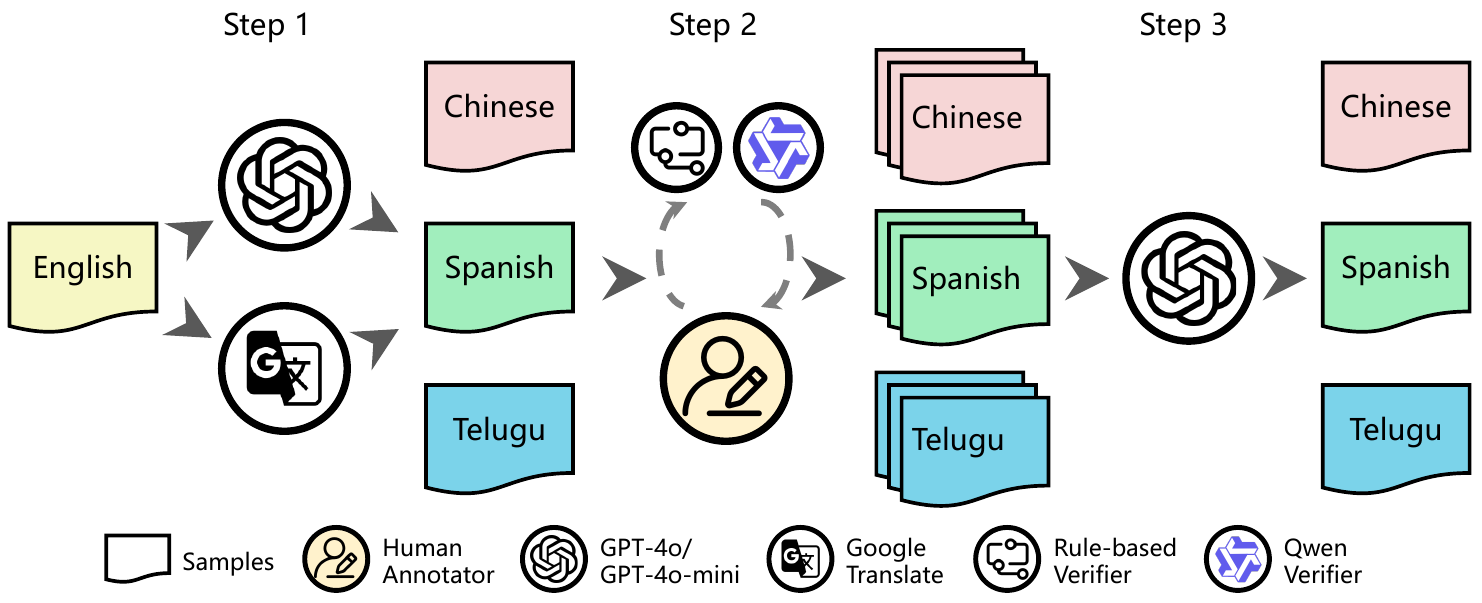}
    \caption{The construction process involves three steps: Step 1) translating data from English to non-English; Step 2) post-editing each sample by three human annotators; Step 3) selecting the final translation version.}
    \label{fig:overview}
    \vskip -0.1in
\end{figure*}

\subsection{Language Selection}
\label{sec:lg_selection}
\name supports 17 selected languages to represent diverse language families and writing systems~(Table~\ref{tab:lg_selection}).

\begin{figure}[t]
    \centering
    \includegraphics[width=0.9\linewidth]{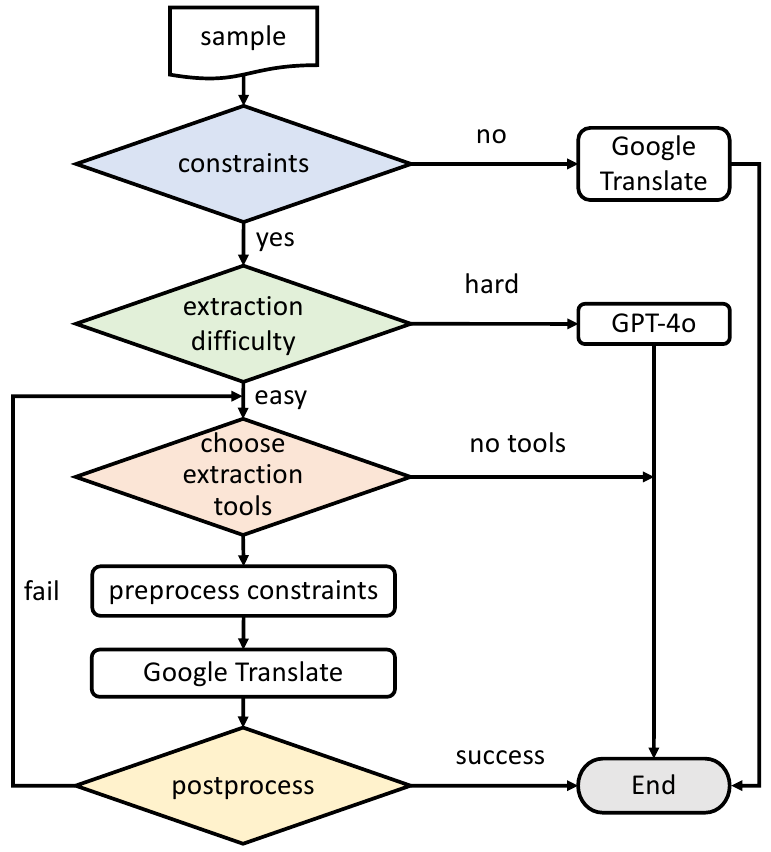}
    \caption{Flow chart illustrating the constraint extraction and machine translation pipeline in the first step of our benchmark construction.}
    \label{fig:step1}
    % \vskip -0.1in
\end{figure}

\subsection{Capabilities Selection}
\label{sec:capability_selection}

LLMs have demonstrated proficiency in understanding tasks such as text classification and sentiment analysis, but their capabilities transcend understanding.
We construct tasks to evaluate following intrinsic capabilities in multilingual settings:
\begin{itemize} [nosep,itemsep=1pt,leftmargin=0.1cm]
    \item \textbf{Instruction Following:} The capability to follow instructions is evaluated by two distinct tasks with different evaluation paradigms: rule-based and model-based assessment. For the rule-based task, we translate IFEval~\cite{zhou2023instruction} from English to other languages, while we expand m-ArenaHard~\cite{dang2024aya,li2024crowdsourced} to languages we selected for the model-based task.
    \item \textbf{Reasoning:} The reasoning capability is assessed through intricate scenarios including math and natural science~(physics, chemistry, and biology) problems. We expand MGSM~\cite{shi2023language} and GPQA~\cite{Rein2023GPQAAG} to 17 languages for the math reasoning and science reasoning tasks.
    \item \textbf{Code Generation:} We primarily consider Python code generation in two settings, function completion and programming problem solving. We translate Humaneval+~\cite{liu2024your,chen2021evaluating} and LiveCodeBench\_v4~\cite{jain2024livecodebench} from English to other languages.
    \item \textbf{Long Context Modeling:} We evaluate the ability to extract evidence from lengthy documents through question-answering tasks with long documents~(128k tokens). We build this task based on RULER~\cite{hsieh2024ruler}, and translate haystacks, needles, and QA pairs.
    \item \textbf{Tool Use:} We assess the ability to correctly select and invoke a single function from multiple available options in response to user queries. We translate the queries in Nexus~\cite{srinivasan2023nexusraven} to other languages, but leave the APIs in English.
    \item \textbf{Translation:} Translation involves accurately converting text between languages while preserving semantic meanings. Beyond traditional translation tasks including Flores, TED, and WMT~\cite{costa2022no,cettolo-etal-2012-wit3,kocmi2024findings}, we introduce the Domain Translation task, a by-product of the \name construction process. This task challenges models to translate specialized terminology and determine whether specific segments should be translated.
\end{itemize}

The information of the seleted datasets, sample sizes, and evaluation metrics is provided in Table~\ref{tab:task_detail}.
More details can be found in Appendix~\ref{sec:appendix-task}.

\renewcommand{\arraystretch}{1.3}
\begin{table*}[t!]
    \centering
    \footnotesize
    % \scalebox{0.9}{
    \begin{tabular}{|p{0.95\linewidth}|}
        \hline
        \textbf{[Original Text]}
        \{prompt: Create an ad copy by expanding ``Get 40 miles per gallon on the highway'' in the form of a QA with a weird style. Your response should contain less than 8 sentences. Do not include keywords `mileage' or `fuel' in your response. \\
        instruction\_id\_list: [`length\_constraints: number\_sentences', `keywords: forbidden\_words'] \\
        kwargs: [\{'relation': `less than', `num\_sentences': 8\}, \{`forbidden\_words': [`mileage', `fuel']\}]\} \\
        \hline
        \textbf{[Translation Input]} 
Create an ad copy by expanding "Get 40 miles per gallon on the highway" in the form of a QA with a weird style. Your response should contain less than 8 sentences. Do not include keywords `\textcolor{red}{$<$b$>$}mileage\textcolor{red}{$<$/b$>$}' or `\textcolor{red}{$<$b$>$}fuel\textcolor{red}{$<$/b$>$}' in your response. \\
        \hline
        
        \textbf{[Google Translation Result]} 
        \begin{CJK}{UTF8}{gbsn}以风格怪异的问答形式扩展“在高速公路上每加仑行驶 40 英里”来创建广告文案。您的回复应少于 8 个句子。请勿在回复中包含关键字“\textcolor{red}{$<$b$>$}里程\textcolor{red}{$<$/b$>$}”或“\textcolor{red}{$<$b$>$}燃料\textcolor{red}{$<$/b$>$}”。\end{CJK} \\

        \hline
    
        \textbf{[Postprocessing]} \{prompt: \begin{CJK}{UTF8}{gbsn}以风格怪异的问答形式扩展“在高速公路上每加仑行驶 40 英里”来创建广告文案。您的回复应少于 8 个句子。请勿在回复中包含关键字“里程”或“燃料”。\end{CJK}
        \\
        instruction\_id\_list: [`length\_constraints: number\_sentences', `keywords: forbidden\_words'] \\
        kwargs: [\{`relation': `less than', `num\_sentences': 8\}, {`forbidden\_words': [\begin{CJK}{UTF8}{gbsn}`里程', `燃料'\end{CJK}]}] \} \\ 
        \hline
        \textbf{[Human Post-Editing]} \{prompt: \begin{CJK}{UTF8}{gbsn}以一种奇特风格的问答形式展开“在高速公路上每加仑行驶40英里”这句话，创建为一个广告文案。你的回答应该少于8句话。不要在你的回复中包含关键字“里程”或“燃料”。\end{CJK}\\ 
        instruction\_id\_list: [`length\_constraints: number\_sentences', `keywords: forbidden\_words']\\
        kwargs: [\{`relation': `less than', `num\_sentences': 8\}, {`forbidden\_words': [\begin{CJK}{UTF8}{gbsn}`里程', `燃料'\end{CJK}]}] \} \\ 
        \hline
    \end{tabular}
    % }
    \caption{One example in rule-based instruction following task, which includes complex constraints. First, we enclose these constraints with special symbols and then translate the prompt from English to the target language by Google Translate. Finally, we postprocess the prompt by extracting constraints into kwargs and removing special symbols for human post-editing.}
    \label{tab:construction-case}
    \vskip -0.1in
\end{table*}

\begin{table}[t]
    \centering
    \footnotesize
    \begin{tabular}{cc|cccc}
        \toprule
         \multicolumn{2}{c|}{\multirow{2}{*}{\textbf{Setting}}} & \multicolumn{4}{c}{\textbf{Target Language}}   \\
         & & zh & es & fr & hu \\
         \midrule
         \multicolumn{2}{c|}{w/o special symbols} & 0.68 & 0.68 & 0.68 & 0.68 
         \\
         \multicolumn{2}{l|}{symbol 1: $<$b$>$ $<$/b$>$} & 0.91 & 0.89 & 0.88 & 0.93 \\
         \multicolumn{2}{l|}{symbol 2: (\texttt{ })} & 0.88 & 0.91 & 0.89 & 0.92 \\
         \multicolumn{2}{l|}{symbol 3: ([\texttt{ }])} & 0.82 & 0.89 & 0.87 & 0.92 \\
        \midrule
         \multirow{2}{*}{Order 1} & + symbol 1 & 0.91 & 0.89 & 0.88 & 0.93 \\
         & + symbol 2 & 0.93 & 0.93 & 0.90 & 0.95 \\
         \midrule
        \multirow{3}{*}{Order 2} & + symbol 2 & 0.88 & 0.91 & 0.89 & 0.92 \\
        & + symbol 1 & 0.90 & 0.93 & 0.90 & 0.95 \\
        & + symbol 3 & 0.92 & 0.93 & 0.90 & 0.95 \\
         \bottomrule
    \end{tabular}
    \caption{The recall rates of constraints using different groups of special symbols. We choose Order 1, which has fewer steps and produces on-par or better performance than other settings.}
    \label{tab:stat_symbol_translation}
    \vskip -0.1in
\end{table}

\subsection{Construction}
\label{sec:construction_process}
The way to obtain \name consists of three steps, as shown in Figure~\ref{fig:overview}: 1) translate data from English to non-English by machines; 2) post-edit each sample by three native annotators; 3) pick the final translation version by GPT-4o-mini.

\paragraph{Step 1: Translating data from English to selected non-English languages by machine translation systems.}
We select between traditional translators such as Google Translate, and LLM-based ones like GPT-4o, depending on whether a task contains extractable constraints.
As illustrated in Figure~\ref{fig:step1}, if the data contains constraints that are hard to extract, we prompt GPT-4o to translate the data and satisfy the constraints.
Otherwise, we use Google Translate along with extraction tools.
Extraction tools include methods for extracting translated keywords by enclosing source keywords with special symbols, and for preserving source constraints by replacing constraints with placeholders before translation and restoring them afterwards.

Taking an example of rule-based instruction following task as an example, as shown in Table~\ref{tab:construction-case}, it requires extra processing to extract constraints from the translated instruction, as they are needed for verification.
Inspired by~\citet{yuan-etal-2020-enhancing}, we enclose the keywords in the original instruction with special symbols, making them easy to extract from the translated result.
If one symbol fails, another symbol is used to improve recall.
As shown in Table~\ref{tab:stat_symbol_translation}, we explore various groups of special symbols and different orders, and calculate the recall rates of keywords.
Comparing to not using special symbols, apply any symbol group can greatly improve the recalls, while combining different symbol groups in multiple rounds can further improve the recalls.
We choose \textit{Order 1} as it can achieve better results with fewer groups than \textit{Order 2}.
The detailed recall results are in Appendix~\ref{sec:append_if_keywords}.

Note that in cases where existing multilingual datasets are available, such as MGSM and m-ArenaHard, we extend them to include the supported languages by translating the English data, to minimize additional effort.

\renewcommand{\arraystretch}{1}
\begin{table*}[t]
    \centering
    \tiny
    \setlength{\tabcolsep}{3.4pt}
    \begin{tabular}{rrcccccccccccc}
    \toprule
        \multirow{3}[2]{*}{\textbf{Model}} & \multirow{3}[2]{*}{\textbf{Size}} & \multicolumn{2}{c}{\textbf{Instruction Following}} & \multicolumn{2}{c}{\textbf{Code Generation}} & \multicolumn{2}{c}{\textbf{Reasoning}} & \textbf{Long Context} & \textbf{Tool Use} & \multicolumn{4}{c}{\textbf{Translation}} \\
        \cmidrule(lr){3-4}\cmidrule(lr){5-6}\cmidrule(lr){7-8}\cmidrule(lr){9-9}\cmidrule(lr){10-10}\cmidrule(lr){11-14}
        & & \multirow{2}{*}{\textbf{Rule-based}} & \multirow{2}{*}{\textbf{Model-based}} & \multirow{2}{*}{\textbf{Func Compl.}} & \multirow{2}{*}{\textbf{Prob. Solving}} & \multirow{2}{*}{\textbf{Math}} & \multirow{2}{*}{\textbf{Science}} & \multirow{2}{*}{\textbf{Question Answering}} & \multirow{2}{*}{\textbf{Multi Func.}} & \multicolumn{2}{c}{\textbf{General}} & \multicolumn{2}{c}{\textbf{Domain}} \\
        & & & & & & & & & & En-X & X-En & En-X & X-En \\
        \midrule
        \multirow{2}{*}{InternLM2.5}     & 7B  & 45.7 & 1.9  & 45.4 & 10.3 & 37.4 & 20.6 & 37.5 & 53.2 & 12.7 & 20.2 & 34.4 & 54.0 \\
             & 20B & 51.9 & 3.3  & 51.2 & 14.4 & 42.9 & 24.0 & -    & 26.6 & 14.9 & 19.7 & 34.9 & 53.9 \\
        \midrule
        \multirow{2}{*}{Aya-Expanse}     & 8B  & 51.2 & 6.4  & 33.8 & 7.8  & 50.8 & 26.2 & -    & 41.1 & 21.5 & 26.8 & 45.6 & 51.6 \\
             & 32B & 61.9 & 12.4 & 52.0 & 15.8 & 66.7 & 27.7 & -    & 59.8 & 25.2 & 32.8 & 54.8 & 62.3 \\
        \midrule
        \multirow{2}{*}{Gemma2}          & 9B  & 63.0 & 9.8  & 53.9 & 16.6 & 72.0 & 23.9 & -    & 61.4 & 27.2 & 33.2 & 57.5 & 61.9 \\
                  & 27B & 62.4 & 18.0 & 66.7 & 24.6 & 75.3 & 26.7 & -    & 64.7 & 30.4 & 34.5 & 64.8 & 66.2 \\
        \midrule
        Llama3.1        & 8B  & 62.6 & 4.3  & 52.9 & 14.1 & 63.4 & 23.8 & 68.3 & 45.0 & 24.6 & 29.8 & 53.9 & 62.9 \\
        R1-Distill-Llama3.1 & 8B & 49.7 & 3.5  & 62.8 & 23.8 & 46.9 & 28.1 & -    & 37.2 & 12.2 & 20.8 & 13.5 & 23.1 \\
        Llama3.1        & 70B & 76.2 & 13.2 & 69.7 & 29.8 & 79.7 & 35.8 & 57.4 & 44.3 & 31.1 & \textbf{35.1} & 64.5 & \textbf{68.2} \\
        Llama3.3                         & 70B & \textbf{85.2} & 17.0 & 74.0 & 34.7 & 83.8 & 42.6 & 50.4 & 42.5 & 31.5 & 33.6 & 63.5 & 65.0 \\
        R1-Distill-Llama3.3 & 70B & 78.0 & 26.6 & \textbf{84.6} & 54.8 & 82.8 & 46.1 & -    & 62.1 & 26.0 & 33.0 & 47.6 & 45.2 \\
        \midrule
        Qwen2.5         & 7B  & 65.9 & 8.5  & 68.2 & 24.7 & 63.4 & 27.6 & 53.5 & 48.9 & 16.6 & 25.6 & 46.4 & 60.0 \\
        R1-Distill-Qwen2.5  & 7B & 46.7 & 3.0  & 69.3 & 37.3 & 56.1 & 28.4 & -    & 27.7 & 6.8  & 16.3 & 17.0 & 27.3 \\
        Qwen2.5         & 32B & 78.1 & 17.3 & 75.8 & 42.7 & 77.7 & 37.7 & 79.4 & 66.7 & 22.7 & 30.5 & 54.2 & 65.4 \\
        R1-Distill-Qwen2.5         & 32B & 67.3 & 19.2 & 80.6 & 54.4 & 77.3 & 37.0 & -    & 60.4 & 20.3 & 28.5 & 37.1 & 37.7 \\
        Qwen2.5         & 72B & 80.8 & 36.9 & 78.6 & 45.5 & 77.8 & 39.4 & 80.6 & 61.8 & 25.8 & 33.3 & 60.4 & 66.9 \\        
        \midrule
        DeepSeek-V3* & 671B & 83.9 & \textbf{59.8} & 83.2 & \textbf{60.4} & \textbf{84.2} & \textbf{47.4} & \textbf{85.2} & 69.2 & \textbf{33.9} & 34.5 & \textbf{70.3} & 67.8 \\
        \midrule
        GPT-4o-mini     & -   & 79.1 & 21.9 & 78.7 & 37.0 & 76.9 & 34.1 & 82.1 & \textbf{70.9} & 30.3 & 33.9 & 67.7 & 67.6 \\
        \bottomrule
    \end{tabular}
    \caption{Performance comparison across models on \name tasks, averaged over 17 languages. Bold numbers indicate the best performance in each column. "Func Compl." refers to Function Completion, "Prob. Solving" to Problem Solving, and "Multi Func." to Multiple Functions scenarios where models must select and call one function from multiple options. Models without results on the long context task do not support 128K context length. * DeepSeek-V3 is a 671B MoE model, with 37B activated for each token.}
    \label{tab:main_results}
    \vskip -0.1in
\end{table*}

\paragraph{Step 2: Post-editing each sample by three distinct native-speaking annotators in almost all tasks.}
To ensure high-quality dataset, we implement a multi-round annotation and verification process.
1) Each sample is given to three native-speaking annotators who are proficient in English and their native language. Considering the specialized nature of datasets like Science reasoning, annotators are required to hold at least a Bachelor's degree. 
2) Two automatic verifiers - rule-based verifiers and model-based verifiers - are used to assess the quality of human annotation. Rule-based verifiers ensure the satisfaction of constraints for certain tasks, such as the rule-based instruction following task.
For model-based verifiers, we utilize the GEMBA-SQM prompt~\cite{kocmi-federmann-2023-large} and employ Qwen2.5-72B-Instruct, a powerful multilingual model, to estimate the quality of translations.
Along with providing an overall score, the model offers detailed explanations of translation errors as feedback to annotators.
Samples that do not pass the rule-based verifier or score below a predefined threshold are identified as failed, and refined in subsequent iterations.
Each manually annotated dataset undergoes at least three iterations.

\paragraph{Step 3: Selecting the final translation version by LLMs.}

Initially, we ask a fourth annotator, uninvolved in the annotation process, to choose the final version from the results revised by three individuals. Intriguingly, the selection by the fourth annotator exhibited a strong position bias, often favoring the initial annotation. This preference could be attributed to the uniformly high quality of translations, resulting in minimal discernible differences among them.

Debiasing for human annotators is costly in terms of both time and finance, because three translations encompass all permutations of six.
Consequently, we employ GPT-4o-mini to select the final translation, as it is a powerful and balanced LLM across different languages.
In particular, following~\citet{li2024crowdsourced}, we adapt the LLM-Judge system instruction~(see Appendix~\ref{sec:prompts}) to suit pairwise translation evaluation.
We shuffle the positions of three translations and conduct two battles to select a final version.
In each battle between two translations, we perform two judgments by swapping their positions and determine one winner.
The winner of the first two translations then battles against the third translation, determining the final winner.

%% file: sections/04_experiments.tex
\section{Experimental Results}

\subsection{Evaluation Setup}

% \vskip 0.1in
\noindent\textbf{Evaluated Models}
We focus primarily on multilingual post-trained models and evaluate both open-source and proprietary language models\footnote{Unless otherwise specified, all models discussed in this paper are post-trained versions.}, including Llama3.1~\cite{dubey2024llama}, Qwen2.5~\cite{qwen2.5}, Gemma2~\cite{team2024gemma}, InternLM2.5~\cite{cai2024internlm2}, Aya-Expanse~\cite{dang2024aya}, DeepSeek-R1-Distill-Llama~\cite{guo2025deepseek}, DeepSeek-R1-Distill-Qwen, DeepSeek-V3~\cite{DeepSeekAI2024DeepSeekV3TR}, and GPT-4o-mini~\cite{openai2024gpt4o}.
Models' detailed descriptions are in Appendix~\ref{sec:model_info}.

% \vskip 0.1in
\noindent\textbf{Inference Configuration}
We adopt greedy decoding for most tasks, except for the problem solving task, where the sampling temperature is set to $0.2$.
The default chat template and system prompt of each model are applied.
% Zero-shot prompts are used for most tasks, while zero-shot CoT prompts are used for reasoning tasks.
Detailed prompts are provided in Appendix~\ref{sec:prompts}.
% Specifically, for instruction following tasks, the translated instructions are directly used as prompts.
For reasoning tasks, we adopt the zero-shot native chain-of-thought templates in LM-Evaluation-Harness~\cite{eval-harness}.
% For the long-context evaluation, the prompt template is in English, while the haystack, needles and questions are in the language being assessed.
% For other tasks, we use the original prompt templates provided in the corresponding repositories\footnote{\url{https://github.com/EleutherAI/lm-evaluation-harness} \\\url{https://github.com/LiveCodeBench/LiveCodeBench} \\\url{https://github.com/evalplus/evalplus} \\\url{https://github.com/NVIDIA/RULER}}, and change the user inputs to other languages.
For other tasks, we use the prompt templates provided in corresponding repositories\footnote{\url{https://github.com/EleutherAI/lm-evaluation-harness} \newline
\url{https://github.com/LiveCodeBench/LiveCodeBench} \newline  
\url{https://github.com/evalplus/evalplus} \newline
\url{https://github.com/NVIDIA/RULER}\newline
\url{https://github.com/lmarena/arena-hard-auto}}, and change the user inputs to other languages.
% More details about the prompt templates can be found in Appendix~\ref{sec:prompts}.

\begin{figure*}[ht]
    \centering
    \includegraphics[width=\linewidth]{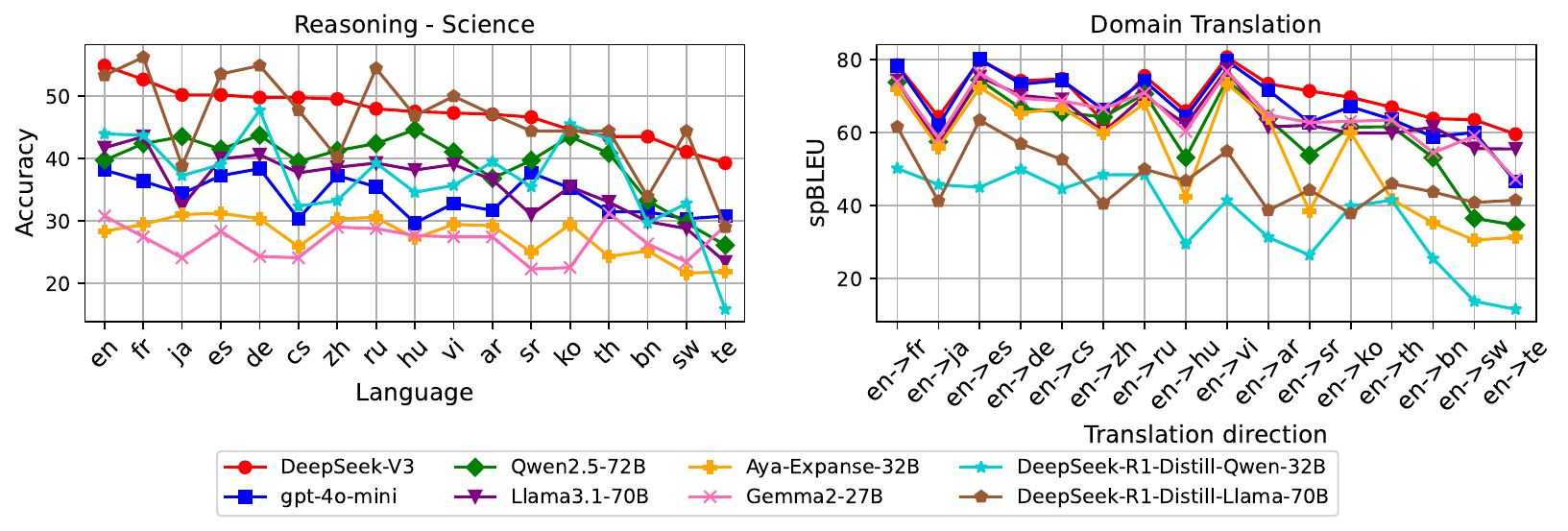}
    \caption{Taking two tasks as examples, models exhibit unbalanced multilingual capabilities. We show performance of several models on the science reasoning task and the domain translation task across different languages.}
    \label{fig:line_plot}
    \vskip -0.1in
\end{figure*}

\begin{figure}[t]
    \centering
    \includegraphics[width=\linewidth]{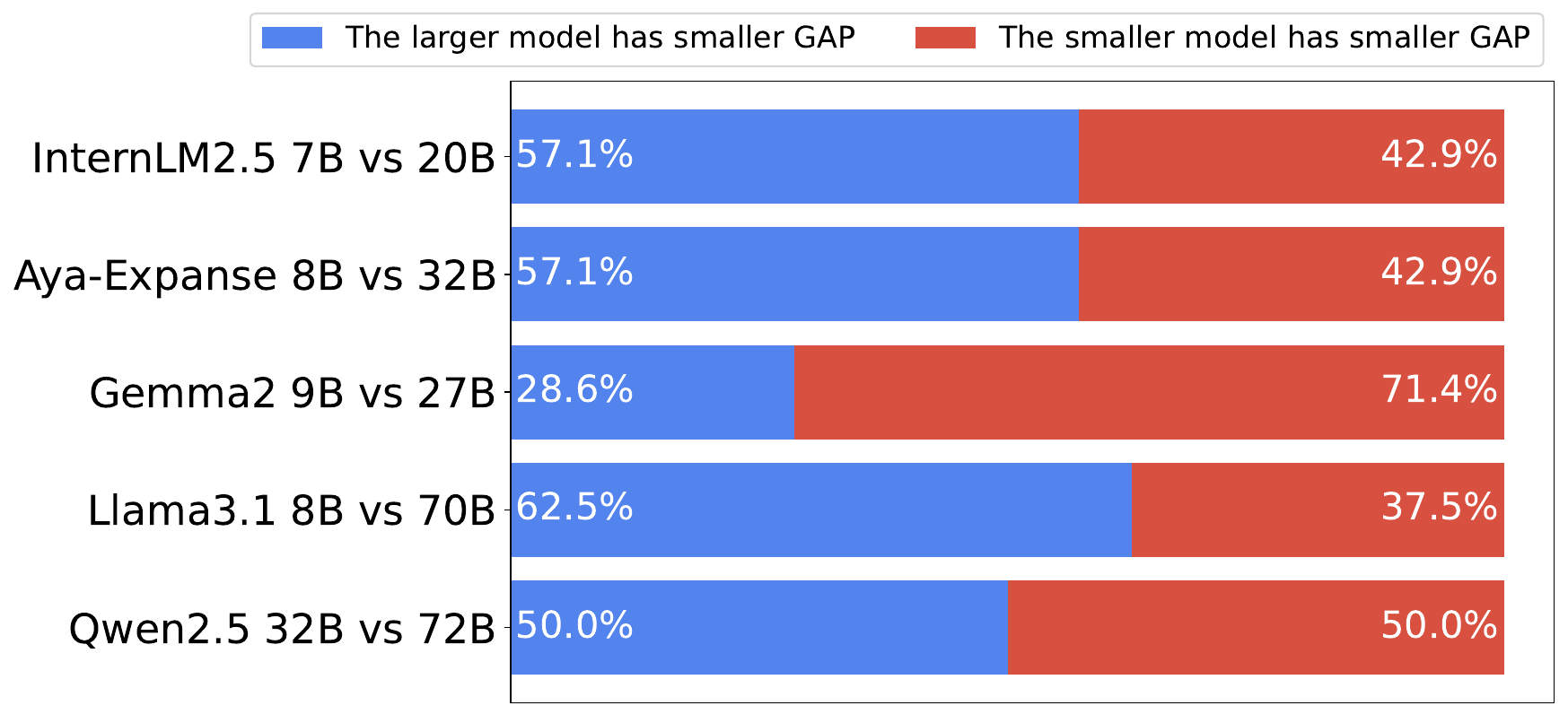}
    \caption{Larger models do not consistently have a smaller GAP. Each row shows proportions of tasks where the larger model achieves a smaller GAP versus where the smaller model performs better.}
    \label{fig:winrate}
    \vskip -0.1in
\end{figure}

\subsection{Multilingual Benchmark Results}
Table~\ref{tab:main_results} shows the overall average performance of each model on each multilingual task.
More detailed results are in Appendix~\ref{sec:detailed_results}.

% Main results are demonstrated in Figure~\ref{fig:radar_chart} and Table~\ref{tab:main_results}.
% Each cell in the table displays the average score computed across 17 languages, along with the average performance gap (GAP) between English and other languages.
% GAP is defined as $\frac{\sum_{l\neq en}\max(s(en) - s(l), 0)}{n-1}$, where $s(l)$ denotes the score on the task with language $l$, and $n$ is the number of languages.

% \paragraph{Scaling the model size enhances the multilingual capabilities consistently, but it does not universally mitigate the performance gap.}

\paragraph{Model scaling improves overall multilingual performance while language disparities persist.}
As shown in Table~\ref{tab:main_results}, larger models consistently demonstrate enhanced multilingual capabilities across all domains, with few exceptions.
% One obvious trend in Table~\ref{tab:main_results} is that as a model becomes bigger, the average multilingual capabilities are enhanced in all domains, with few exceptions.
% All evaluated capabilities of Qwen2.5-72B, surpass those of Qwen2.5-7B in multilingual settings. % including instruction following, code generation, reasoning and so on
However, the performance gap between English and non-English languages does not invariably diminish.
We define GAP as the average performance gap between English and other languages:
$$GAP=\frac{\sum_{l\neq \mathrm{en}}\max(s(\mathrm{en}) - s(l), 0)}{n-1},$$ where $s(l)$ denotes the score on the task with language $l$, and $n$ is the number of languages including English.
As shown in Figure~\ref{fig:winrate}, when comparing models of different sizes, the proportion of larger models achieving smaller GAPs only slightly exceeds 0.5 for most model families.
Gemma2-9B achieves smaller GAPs than Gemma2-27B on most tasks.
These findings suggest that while scaling model size effectively improves overall multilingual performance, additional strategies may be needed to address the performance disparities across languages.
% This indicates that simply scaling the model size is not a promising way to completely close the performance gap between English and other languages.
% Gemma2-9B achieves even smaller GAPs than Gemma-27B in most tasks.
% As depicted in Figure~\ref{fig:winrate}, most proportions of larger models with smaller GAPs, except for Qwen2.5, are just slightly above 0.5.
% We find that as the model size increases, GAPs on most tasks decreases, but there are still a number of exceptions.
% As demonstrated in Figure~\ref{fig:radar_chart} which shows detailed results of each task in different languages, the capability enhancement of Llama3.1 and Qwen2.5 can be observed in all languages when the average task performance is improved, suggesting that scaling model size can consistently improve the multilingual capabilities across all languages.
% Qwen2.5 shows an improvement of at least 11\% on the rule-based instruction following task across all evaluated languages.
% However, Llama3.1 does not exhibit improvement on the long-context modelling task and the tool use task, which may attributed to the limitations in the corresponding capabilites of Llama3.1-70B.

% \vskip 0.1in
\paragraph{The effective utilization of language-agnostic capabilities remains challenging in multilingual contexts.}
As illustrated in the left plot of Figure~\ref{fig:line_plot}, models' reasoning capabilities vary significantly across languages.
Generally, models achieve better performance in dominant languages like English than in other low-resource languages.
This disparity can be attributed to the fact that multilingual task execution depends not only on language-agnostic reasoning but also on language-specific capabilities such as comprehension and generation.
Therefore, when operating in weak languages, it becomes difficult for a model to fully leverage its language-agnostic capabilities.
Interestingly, we observe an unexpected pattern where certain models excel in specific non-dominant languages compared to English on some tasks.
For example, Qwen2.5 demonstrates superior performance in Korean over English on the science reasoning task.
We hypothesize that Qwen2.5’s training data includes a relatively high proportion of Korean content in scientific or reasoning-related domains.
This counter-intuitive phenomenon warrants further investigation in future work.

\begin{figure*}[ht]
    \centering
    \includegraphics[width=0.9\linewidth]{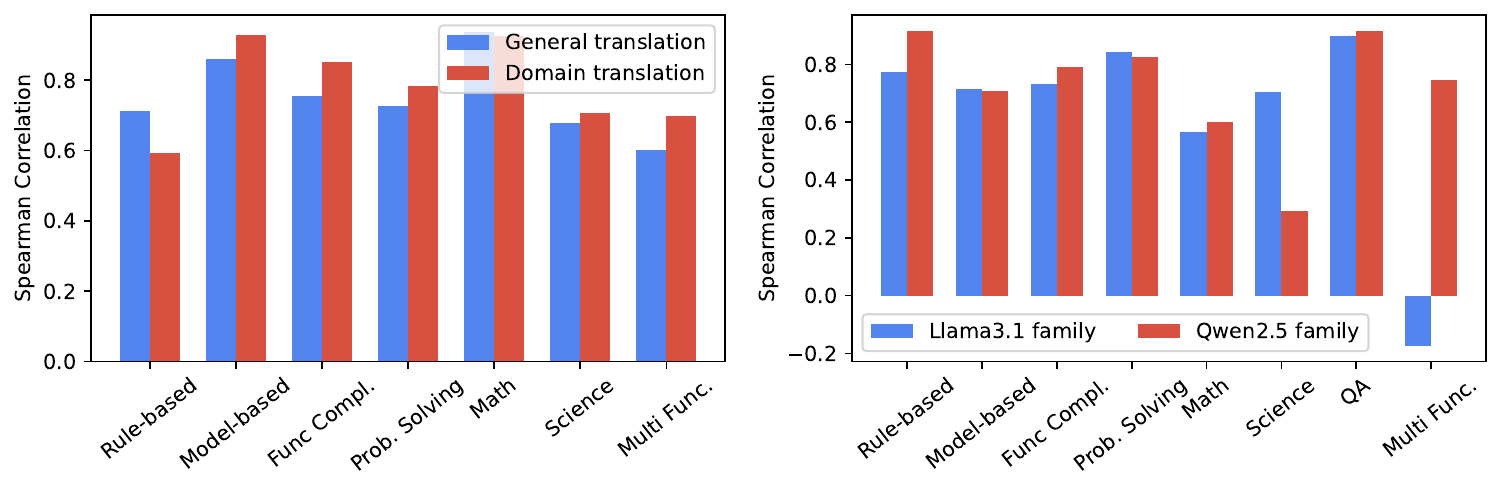}
    \caption{\textbf{Left:} The translation performance is positively correlated with other multilingual performance. Spearman Correlations are computed between the performance on general/domain translation and the specific task. \textbf{Right:} Models in the same family have similar language performance pattern. We compute the Spearman Correlations between the performance of two models (Llama3.1 8B vs 70B, Qwen2.5 7B vs 72B) across different languages.}
    \label{fig:spearman}
    \vskip -0.1in
\end{figure*}

\paragraph{Model performance exhibits systematic bias towards high-resource languages.}
As shown in Figure~\ref{fig:line_plot}, the performance curves of most models exhibit significant fluctuations across languages.
High-resource languages such as French and Chinese consistently outperform low-resource languages like Telugu, Swahili, and Bengali.
For instance, while DeepSeek-V3 achieves over 50\% accuracy in science reasoning tasks for English and French, its performance drops notably to below 40\% for Telugu.
% Our evaluation shows that almost all models have at least one language where they significantly underperform.
This pattern can be partially attributed to development strategies - models like Aya-Expanse were not specifically optimized for the full range of languages in our evaluation.
Unexpectedly, Gemma2 exhibits relatively balanced performance across most tasks (Figure~\ref{fig:radar_chart}), despite not being explicitly marketed as a multilingual model.

% \vskip 0.1in
\paragraph{Translation capabilities exhibit a positive correlation with other evaluated capabilities.}
We analyze the relationship between model's English-to-X translation capability and other capabilities using Spearman correlation coefficients (the left panel of Figure~\ref{fig:spearman}).
When calculating correlations between domain-specific translation performance and task performance, we exclusively use data from the corresponding domains.
The analysis reveals that domain-specific translation performance generally exhibits stronger correlations with task performance compared to general translation capabilities.
A notable exception is that in the rule-based instruction-following task, we observe an inverse scaling effect: larger LLMs produce lower-quality translations compared to their smaller counterparts.
We find that larger LLMs are more likely to execute instructions rather than strictly perform translation, known as prompt injection.

\paragraph{Models within the same family exhibit consistent performance patterns across languages.}
We calculate Spearman correlation coefficients to analyze the performance similarity between models of the same family (excluding R1-distilled models) across different languages for each task.
As shown in the right panel of Figure~\ref{fig:spearman}, models within the same family show strong correlations across various tasks, with most correlation coefficients exceeding 0.7.
% The exception is Llama3.1's negative correlation in tool use tasks.
% More results on other tasks are illustrated in Figure~\ref{fig:radar_chart}, where we can find that Llama3.1-8B and Llama3.1-70B exhibit similar performance pattern across various tasks.
% Both Qwen2.5-7B and Qwen2.5-72B perform relatively poor on Swahili and Telegu on all tasks.

\paragraph{R1-distilled models exhibit enhanced multilingual reasoning and code generation capabilities, but some other capabilities, especially translation, are noticeably degraded.}
As illustrated in Table~\ref{tab:main_results}, the performance of R1-Distill-Llama3.3-70B is comparable to DeepSeek-V3 in reasoning and code generation tasks, and is stronger than Llama-3.3-70B-Instruct.
However, other capabilities like instruction following of 7B/8B models exhibit degradation to some extent.
They tend to generate repeated tokens in the reasoning process when using non-English languages.
The translation capabilities of both large and small distilled models decline dramatically.
In addition to repeated generation, we also observe a frequent phenomenon of code-switching in translations.

%% file: sections/05_analysis.tex
\section{Analysis}

\begin{table}[t]
    \centering
    \footnotesize
    \setlength{\tabcolsep}{3pt}
    \begin{tabular}{l|ccc|ccc}
    \toprule
         & \multicolumn{3}{c|}{Llama3.1-70B} & \multicolumn{3}{c}{Qwen2.5-72B} \\
        Translated by & GT & 4o-mini & Ours & GT & 4o-mini & Ours \\
    \midrule
        Rule-based      & 66.9 & 53.5 & \textbf{76.2} & 71.5 & 57.2 & \textbf{80.8} \\
        Func Compl.     & 47.8 & 68.2 & \textbf{69.7} & 50.4 & 75.5 & \textbf{78.6} \\
        Science         & 33.7 & 35.1 & \textbf{35.8} & 36.9 & 37.8 & \textbf{39.4} \\
        Multi Func.     & 23.0 & 43.7 & \textbf{44.3} & 26.7 & 61.3 & \textbf{61.8} \\
    \bottomrule
    \end{tabular}
    \caption{Our pipeline provides a more accurate assessment of the multilingual performance, compared to naive translations by Google Translate(GT) and GPT-4o-mini(4o-mini), respectively.}
    \label{tab:naive_results}
    \vskip -0.1in
\end{table}

% \subsection{Naively machine-translated task data underestimate LLMs' capabilities}
\subsection{Our pipeline provides a more accurate assessment of models' performance}
We naively translate a subset of tasks from English to other 16 languages by Google Translate and GPT-4o-mini, and then evaluate two models using this task data.
We directly translate sources by Google Translate as it doesn't support constraints, and use appropriate prompts with constraints to request GPT-4o-mini.
The results in Table~\ref{tab:naive_results} show that models achieve higher scores generally on our translated tasks compared to naive machine-translated ones.
Google Translate lacks the flexibility to handle diverse constraints and specific domains, while GPT-4o-mini does not always perform translation task, especially on instruction data.
This indicates that naive machine translation underestimates LLMs' capabilities, whereas our data provides a more accurate assessment.
% Our pipeline can take advantage of both translation models and LLMs, providing a more accurate assessment of the true performance.

\subsection{\name is adequate to evaluate multilingual capabilities of LLMs}
Prior work like Aya-Expanse~\cite{dang2024aya} relies on conventional understanding tasks such as XCOPA and XWinograd for multilingual evaluation.
On these metrics, Gemma2-9B achieves the best performance, followed by Aya-Expanse-8B, Llama3.1-8B, and Qwen2.5-7B.
However, our evaluation through \name reveals a different pattern: Qwen2.5-7B demonstrates superior multilingual capabilities on generation tasks, while Aya-Expanse models show notably weaker performance on code generation tasks, as shown in Table~\ref{tab:aya_task}.
This discrepancy highlights the importance of comprehensive evaluation frameworks that incorporate both understanding and generation tasks to accurately assess multilingual capabilities of LLMs.

\begin{table}[!t]
    \centering
    \scriptsize
    \begin{tabular}{c|cc|c}
    \toprule
        \textbf{Model} & \textbf{Rule-based} & \textbf{Func Compl.} & \textbf{Discriminative} \\
    \midrule
        Qwen2.5-7B & No.1 & No.1 & No.4 \\
        Llama3.1-8B & No.3 & No.3 & No.3 \\
        Aya-Expanse-8B & No.4 & No.4 & No.2 \\
        Gemma2-9B & No.2 & No.2 & No.1 \\
    \bottomrule
    \end{tabular}
    \caption{Rankings of the models in generative tasks in \name differ from that in discriminative tasks, indicating the importance of both types of tasks. The ranking in discriminative tasks is from \citet{dang2024aya}.}
    \label{tab:aya_task}
    \vskip -0.1in
\end{table}

\subsection{\name reveals the challenges in domain-specific translation evaluation}
% \subsection{Comparison of domain translation metrics}
Domain-specific texts often contain substantial content, such as code, that does not require translation, leading to inflated spBLEU scores.
% A substantial proportion of domain-specific text does not require translation, leading to extremely high spBLEU.
To address this limitation, we explore alternative evaluation approaches: the edit-distance metric TER~\cite{snover2006study}, the model-based metric \textsc{xCOMET}-XXL~\cite{guerreiro2024xcomet}, and a novel performance retention rate that compares downstream task performance between model self-translation and human translations.
% We also consider other machine translation metrics, including an edit-distance metric, TER~\cite{snover2006study}, and a model-based metric \textsc{xCOMET}-XXL~\cite{guerreiro2024xcomet}.
% Another potential approach is to utilize the translations as downstream task data and then evaluate a model on this data.
% The retention rate of performance on model tanslations compared to that on human translations can be used as a metric for translation quality.
Table~\ref{tab:domain_metrics} presents these metric scores across selected tasks and languages. % , revealing several evaluation challenges.
Traditional metrics prove unreliable for domain-specific translation evaluation.
Both spBLEU and TER show extreme values in Science and Programming tasks due to large portions of unchanged text, failing to capture the quality of crucial translated segments.
The model-based metric \textsc{xCOMET} scores show inconsistency, from 18 to 96, across different scenarios, particularly struggling with low-resource languages and specialized domains.
Moreover, the performance retention rate doesn't work as expected as many values are very close, which fails to evaluate translations directly and subtly.
These findings highlight the need for specialized metrics for domain-specific translation evaluation, which we identify as an important direction for future research.

\begin{figure}[t]
    \centering
    \includegraphics[width=\linewidth]{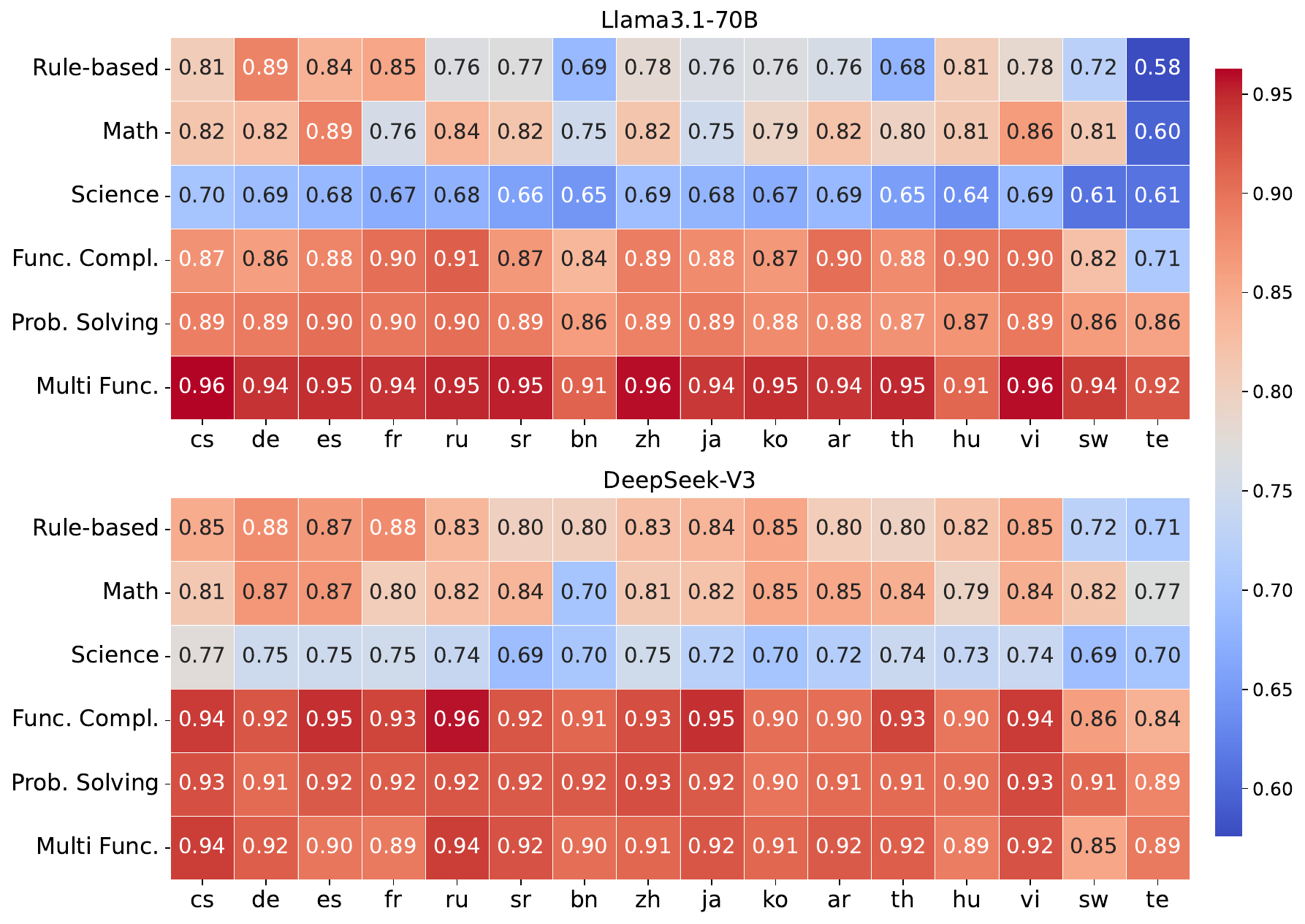}
    \caption{Advanced models show high consistency between English and other languages across six tasks.}
    \label{fig:agreement}
    \vskip -0.1in
\end{figure}

\subsection{High consistency between the questions answered correctly/incorrectly in English and in other languages}
Although sometimes similar performance can be achieved across different languages for certain tasks, the specific problems being addressed may vary significantly.
To examine the language alignment, we compute the consistency between the problem-solving correctness in English versus other languages.
Consistency is calculated as the proportion of predictions where a model's output is correct or incorrect in both languages, out of all evaluated samples.
Figure~\ref{fig:agreement} presents the consistency between English and languages, based on results of Llama3.1-70B and DeepSeek-V3 on six subtasks of \name.
Both these strong multilingual models demonstrate high consistency on most tasks, with most scores exceeding 0.75.
Agreement for low-resource languages are notably lower than those for high-resource languages.
Low consistency is also pronounced in science reasoning tasks, suggesting these knowledge-intensive problems pose unique challenges for cross-lingual knowledge transfer.
% To examine the extent of correctness agreement, we compute F1 scores measuring the agreement between the correctness of each problem in English and its corresponding correctness in another language.
% The correctness in English is treated as labels.
% Table~\ref{fig:agreement} demonstrates the F1 scores of Llama3.1-70B and Qwen2.5-72B on five subtasks of \name between English and other languages.
% Both the two strong multilingual models have a high level of correctness agreement on most tasks, with average F1 scores exceeding 0.9.
% As anticipated, F1 scores on the low-resource languages are comparatively lower those on high-resource languages.
% The relative low agreement on the science reasoning knowledge-intensive task may indicate that knowledge is difficult to share between languages.

% We find that consistency is lower for low-resource languages than for high-resource ones, which is similar to the pattern of average performance.
% Furthermore, the consistency observed in the last three tasks is significantly higher compared to the first two tasks.
% We hypothesize that demanding more language-specific capabilities leads to lower consistency, where models are required to generate native CoT in the reasoning tasks.
% In contrast, the subsequent three tasks primarily require models to generate outputs in programming languages or English function calls, which do not challange the multilingual generation capabilities.

\begin{table*}[ht]
    \scriptsize
    \centering
    \begin{tabular}{c|c|cccc|cccc|cccc}
        \toprule
        \multirow{2}{*}{\textbf{Metric}} & \multirow{2}{*}{\begin{tabular}[c]{@{}c@{}}\textbf{Translation}\\ \textbf{Model}\end{tabular}} & \multicolumn{4}{c|}{\textbf{Reasoning - Math}} & \multicolumn{4}{c|}{\textbf{Reasoning - Science}} & \multicolumn{4}{c}{\textbf{Code generation - Prob. Solving}} \\
         & & zh & de & sw & te & zh & de & sw & te & zh & de & sw & te \\
        \midrule
        \multirow{3}{*}{spBLEU}& Gemma2-27B & 40.0 & 51.4 & 38.2 & 29.2 & 80.6 & 84.8 & 66.2 & 57.5 & 85.5 & 78.5 & 76.2 & 52.3 \\
        & Llama3.1-70B & 35.2 & 54.4 & 36.6 & 35.0 & 71.8 & 84.9 & 64.0 & 65.3 & 84.8 & 78.5 & 75.5 & 56.7 \\
        & Qwen2.5-72B & 37.7 & 50.1 & 13.5 & 12.6 & 77.0 & 79.4 & 41.2 & 40.9 & 84.6 & 73.0 & 48.8 & 42.1 \\
        \midrule
        \multirow{3}{*}{TER} & Gemma2-27B & 36.2 & 32.1 & 40.2 & 58.6 & 15.7 & 12.8 & 26.9 & 33.5 & 15.3 & 15.6 & 17.4 & 33.3 \\
        & Llama3.1-70B & 36.0 & 30.1 & 44.0 & 51.8 & 19.6 & 12.9 & 28.4 & 26.9 & 15.1 & 15.4 & 17.9 & 46.8 \\
        & Qwen2.5-72B & 33.1 & 33.5 & 76.8 & 85.7 & 15.4 & 16.8 & 65.8 & 51.4 & 14.3 & 19.6 & 38.3 & 53.5 \\
        \midrule
        \multirow{3}{*}{\textsc{xCOMET}} & Gemma2-27B & 86.0 & 96.1 & 68.1 & 71.8 & 63.2 & 77.6 & 36.3 & 44.1 & 45.2 & 46.7 & 27.0 & 25.0 \\
        & Llama3.1-70B & 86.8 & 95.6 & 66.1 & 74.0 & 63.7 & 77.4 & 37.1 & 46.8 & 43.5 & 46.3 & 27.8 & 28.9 \\
        & Qwen2.5-72B & 87.6 & 95.6 & 24.5 & 30.1 & 65.2 & 76.3 & 20.3 & 28.4 & 45.0 & 45.7 & 18.0 & 18.4 \\
        \midrule
        \multirow{3}{*}{Retention Rate} & Gemma2-27B & 1.00 & 1.08 & 0.98 & 0.99 & 0.98 & 0.98 & 1.07 & 0.81 & 1.02 & 0.96 & 0.89 & 0.95 \\
        & Llama3.1-70B & 1.01 & 1.06 & 1.00 & 0.97 & 0.92 & 1.06 & 0.99 & 0.77 & 1.01 & 0.98 & 0.91 & 0.89 \\
        & Qwen2.5-72B & 1.03 & 1.04 & 0.71 & 0.71 & 0.90 & 0.98 & 1.04 & 0.79 & 1.00 & 1.04 & 0.93 & 0.86 \\
        \bottomrule
    \end{tabular}
    \caption{There exists challenges in domain-specific translation evaluation. The table presents different metric scores of the En-X translation of selected models on specific domains.}
    \label{tab:domain_metrics}
    \vskip -0.1in
\end{table*}

\subsection{Self-bias is inevitable in the model-based instruction following tasks}
\begin{figure}[t]
    \centering
    \includegraphics[width=\linewidth]{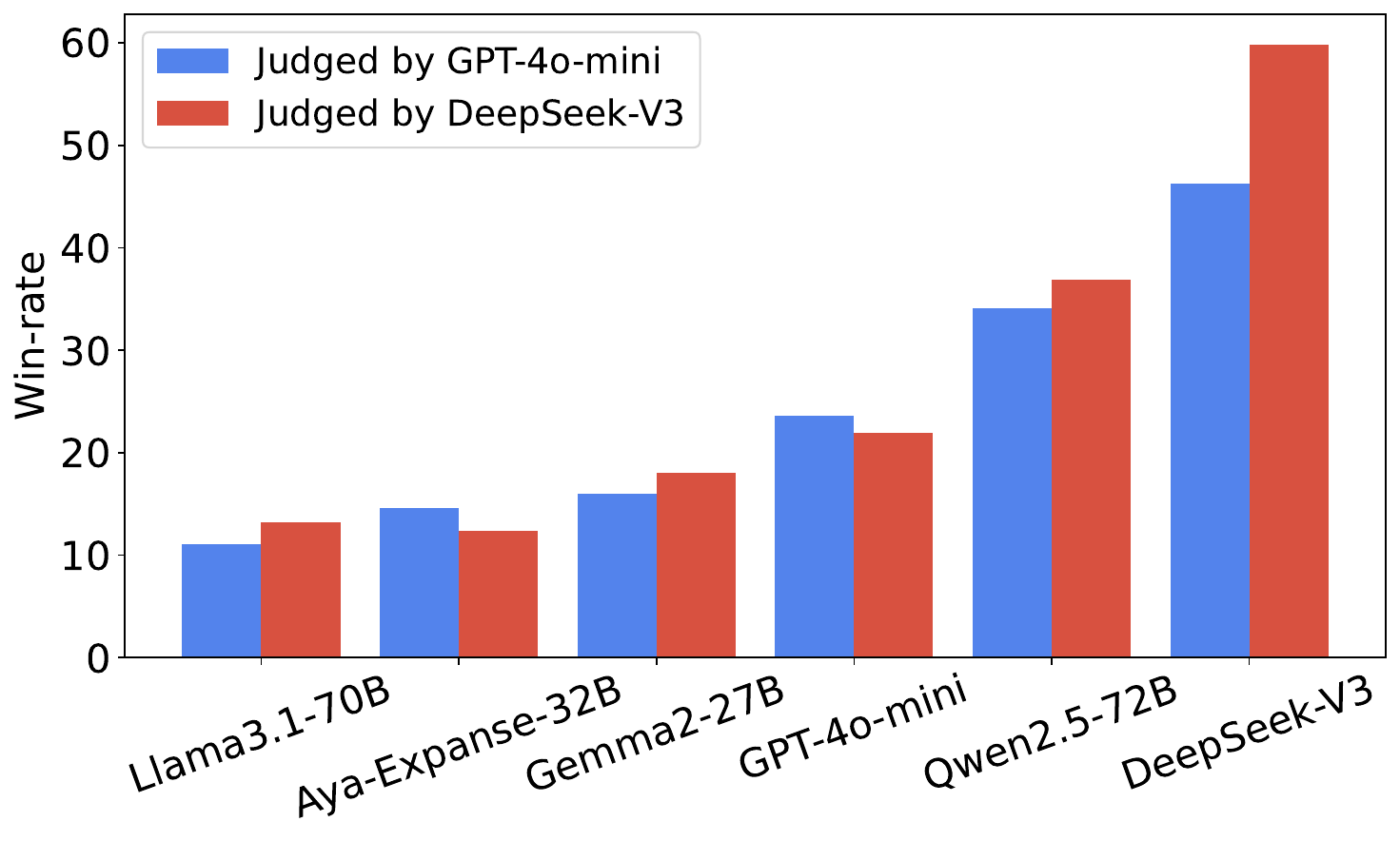}
    \caption{Self-bias is inevitable in model-based instruction following evaluation. DeepSeek-V3 prefers its own outputs, while GPT-4o-mini tends to prefer GPT-4o's outputs. The win-rates of evaluated models are judged by DeepSeek-V3 and GPT-4o-mini.}
    \label{fig:selfbias}
    \vskip -0.1in
\end{figure}

Applying model-based evaluation exhibits self-bias, where the judge model prefers the outputs of itself or models from the same family~\cite{li2024crowdsourced, xu-etal-2024-pride}.
We further adopt GPT-4o-mini as the judge model, and compute the win-rate against the baseline model GPT-4o.
Figure~\ref{fig:selfbias} shows that DeepSeek-V3 strongly favors its own outputs, while GPT-4o-mini prefers GPT-4o's outputs.
Nevertheless, the win-rates of other models judged by the different judges are comparable, and the rankings are fairly consistent.

\subsection{Comaparing open-source and closed-source models on \name}
As demonstrated in Table~\ref{tab:4o_result}, although GPT-4o-mini and GPT-4o demonstrates strong multilingual capabilities across various tasks, they fall short of DeepSeek-V3, the leading open-source model in our evaluation.
This suggests that state-of-the-art open-source models are becoming competitive with their closed-source counterparts.
Due to budget constraints, our evaluation of closed-source models is limited to GPT-4o-mini and GPT-4o on some of tasks.
A more comprehensive comparison would be valuable for further validating this trend.
% However, Due to limited budget, our evaluation of closed-source models focuses solely on GPT-4o-mini. a more comprehensive comparison involving additional closed-source models would be valuable for validating this trend.
% \paragraph{Comprison between open-source models and closed-source models}
% We only evaluate GPT-4o-mini on our benchmarks due to limited resources.
% As demonstrated in Table~\ref{tab:main_results} and Figure~\ref{fig:radar_chart}, GPT-4o-mini exhibits strong multilingual capabilities across various tasks.
% Its performance on several tasks is comparable to that of the open-source model Qwen2.5-72B, and inferior to the strongest open-source model evaluated in this study, DeepSeek-V3.
% We believe that, with the emergence of DeepSeek-V3, the gap between open-source and closed-source models is narrowing.
% Further efforts are required to conduct a comprehensive comparison between open-source and closed-source models.

\begin{table}[!t]
    \centering
    \small
    \setlength{\tabcolsep}{4pt}
    \begin{tabular}{c|ccc}
    \toprule
        \textbf{Model} & \textbf{Rule-based} & \textbf{Func Compl.} & \textbf{Science} \\
    \midrule
        GPT-4o-mini & 79.1 & 78.7 & 34.1 \\
        GPT-4o      & 80.8 & 80.1 & 45.6 \\
        DeepSeek-V3 & \textbf{83.9} & \textbf{83.2} & \textbf{47.4} \\
    \bottomrule
    \end{tabular}
    \caption{The leading open-source model, DeepSeek-V3, bridges the gap to the closed-source models. We compare DeepSeek-V3 and GPT-4o on some of tasks.}
    \label{tab:4o_result}
    \vskip -0.1in
\end{table}

%% file: sections/06_conclusion.tex
\section{Conclusion}
In this work, we introduce \name, a comprehensive, high-quality, and parallel multilingual benchmark comprising ten tasks designed to assess crucial capabilities across seventeen diverse languages.
The multilingual task data is initially translated from English using machine translation and subsequently refined through multiple iterations of post-editing by native speakers, ensuring high data quality.
Through extensive model evaluations, we find that the language-agnostic capabilities of state-of-the-art LLMs remain uneven across different languages.
While increasing model size consistently enhances multilingual performance, the performance gap between English and other languages persists, highlighting the need for further efforts to achieve balanced multilingual capabilities.

% \section*{Impact Statement}
% This paper presents work whose goal is to advance the field of 
% Machine Learning. There are many potential societal consequences 
% of our work, none which we feel must be specifically highlighted here.

% \section*{Acknowledgments}
% 
% This document has been adapted
% by Steven Bethard, Ryan Cotterell and Rui Yan
% from the instructions for earlier ACL and NAACL proceedings, including those for
% ACL 2019 by Douwe Kiela and Ivan Vuli\'{c},
% NAACL 2019 by Stephanie Lukin and Alla Roskovskaya,
% ACL 2018 by Shay Cohen, Kevin Gimpel, and Wei Lu,
% NAACL 2018 by Margaret Mitchell and Stephanie Lukin,
% Bib\TeX{} suggestions for (NA)ACL 2017/2018 from Jason Eisner,
% ACL 2017 by Dan Gildea and Min-Yen Kan,
% NAACL 2017 by Margaret Mitchell,
% ACL 2012 by Maggie Li and Michael White,
% ACL 2010 by Jing-Shin Chang and Philipp Koehn,
% ACL 2008 by Johanna D. Moore, Simone Teufel, James Allan, and Sadaoki Furui,
% ACL 2005 by Hwee Tou Ng and Kemal Oflazer,
% ACL 2002 by Eugene Charniak and Dekang Lin,
% and earlier ACL and EACL formats written by several people, including
% John Chen, Henry S. Thompson and Donald Walker.
% Additional elements were taken from the formatting instructions of the \emph{International Joint Conference on Artificial Intelligence} and the \emph{Conference on Computer Vision and Pattern Recognition}.

%% file: sections/07_appendix.tex
\clearpage
\appendix

\section{Capability and Task Data Selection}
\label{sec:appendix-task}

% \subsection{Capability Selection}

\paragraph{Instruction Following Capability} involves understanding and executing commands accurately and efficiently. In the light of varied evaluation methods - rule-based or model-based - we include two distinct tasks.

\begin{itemize}[nosep,itemsep=1pt,leftmargin=0.1cm]
    \item \underline{Rule-based Intruction Following:} We collect data from IFEval~\cite{zhou2023instruction}, which is a benchmark for evaluating the instruction following abilities of LLMs, composed of around 500 verifiable instructions and can be evaluated for accuracy using automated rules.
    % After filtering out all English-specific instructions, such as changing the letter cases, the number of remaining samples is 429.
    Note that the accuracy for IFEval is the average of the four accuracies~(i.e. prompt-strict, prompt-loose, inst-strict and inst-loose accuracies), following \cite{dubey2024llama}.
    \item \underline{Model-based Instruction Following:} We collect data from Arena-hard~\cite{li2024crowdsourced} which contains 500 real-world instructions from the Chatbot Arena~\cite{chiang2024chatbot}, and m-ArenaHard\footnote{\url{https://huggingface.co/datasets/CohereForAI/m-ArenaHard}} which contains translated multilingual versions. This benchmark can provide better model separability and higher alignment with human preference. It is assessed by the Win Rate of the testing model in comparison to the baseline model, GPT-4o, judged against DeepSeek-V3.
\end{itemize}

\paragraph{Code Generation Capability} refers to automatically producing functional code scripts based on given requirements. Considering variations in difficulty, two separate tasks are included. 
\begin{itemize}[nosep,itemsep=1pt,leftmargin=0.1cm]
    \item \underline{Function Completion:} We collect data from Humaneval+~\cite{liu2024your} which is an augmented version of HumanEval~\cite{chen2021evaluating}, comprising an expanded test cases. Each problem in the benchmark gives a definition of a Python function accompanied by an English docstring, and requires LLMs to complete the function.
    \item \underline{Problem Solving:} We collect data from LiveCodeBench~\footnote{We adopt the code generation subset in LiveCodeBench v4 as the original English dataset.}~\cite{jain2024livecodebench} which provides a more rigorous assessment of the code generation capabilities. It is a much harder benchmark by collecting coding problems in natural language from real competition platforms with live updates.
\end{itemize}

\paragraph{Long Context Modeling Capability} involves understanding and generating coherent text from extensive input sequences, allowing the model to capture dependencies and relationships within lengthy texts. This paper focuses on the long-context evaluation of multilingual settings based on the RULER benchmark~\cite{hsieh2024ruler}.
\begin{itemize}[nosep,itemsep=1pt,leftmargin=0.1cm]
\item \underline{Question Answering:} We build synthetic testsets based on RULER, which contains several question answering long-context tasks with pre-defined context length, such as the needle-in-a-haystack~(NIAH) test and question answering~(QA) test.
Since the NIAH test is unrealistic and many models perform perfectly on it, we add a new task called QA-in-a-heystack~(QAIAH), where one or several paragraphs are inserted into the haystack. The model then answers the question related to the inserted paragraph instead of finding the obtrusive needle. We reserve the tasks of NIAH, QAIAH, and variable tracking~(VT) in our task list, while others are excluded. 
\end{itemize}

\paragraph{Reasoning} encompasses thinking logically, drawing conclusions, making inferences, and solving problems by processing data, applying rules, and utilizing various forms of logic and knowledge representation. Pushing LLMs beyond surface-level tasks, we extend MGSM~\cite{shi2023language} and GPQA~\cite{Rein2023GPQAAG} requiring deeper understanding and reasoning across different context.
\begin{itemize}[nosep,itemsep=1pt,leftmargin=0.1cm]
    \item \underline{Math Reasoning:} We collect data from MGSM which evaluates the capability of LLM to solve math reasoning problems in multiple languages, focusing on grade-school level complexity.
    \item \underline{Science Reasoning:} We collect data from GPQA which is crucial for assessing LLM capability for advanced, unsearchable reasoning and critical thinking across diverse, complex domains. It comprises multiple choice questions formulated by experts in the domains of biology, physics, and chemistry, posing extreme challenges where human experts achieve accuracy lower than 70\%.
\end{itemize}

\paragraph{Tool Use Capability} requires the model to translate user queries into executable functions for calling in operating software tools. We extend Nexus~\cite{srinivasan2023nexusraven} to a multilingual version, which is adopted by Llama3~\cite{dubey2024llama}. 
\begin{itemize}[nosep,itemsep=1pt,leftmargin=0.1cm]
    \item \underline{Multiple Functions:} Nexus offers a set of functions and user queries. For each query, the language model is required to generate a function call from a list of noisy functions, in accordance with the function definitions and docstrings. 
\end{itemize}

\paragraph{Translation Capability} needs the model to convert text between multiple languages while maintaining semantic meaning accurately. To comprehensively evaluate this capability, we introduce general and task-specific translation datasets.
\begin{itemize}[nosep,itemsep=1pt,leftmargin=0.1cm]
    \item \underline{General:} General domain data are composed of Flores-200~\cite{costa2022no}, TED~\cite{cettolo-etal-2012-wit3} and WMT24~\cite{kocmi2024findings} testsets. In \name, we include parallel data from 17 selected languages.
    \item \underline{Domain:} Domain translation data is a by-product of the \name construction process, encompassing a 17-way parallel task across diverse domains, such as reasoning, code generation, tool usage, and instruction following. Unlike traditional translation tasks, this poses a new challenge to the model by requiring it to determine whether a given segment should be translated or not.
\end{itemize}

\begin{table*}[!t]
    \centering
    \small
    \setlength{\tabcolsep}{3pt}
    \begin{tabular}{ccccccccccccccccc}
    \toprule
    & zh & es & fr & de & hu & ru & ja & th & sw & bn & te & ar & ko & vi & cs & sr \\
    \midrule
    w/o special symbols & 0.68 & 0.68 & 0.68 & 0.68 & 0.68 & 0.68 & 0.68 & 0.68 & 0.68 & 0.68 & 0.68 & 0.68 & 0.68 & 0.68 & 0.68 & 0.68 \\
    +symbol 1 & 0.91 & 0.89 & 0.88 & 0.92 & 0.93 & 0.93 & 0.91 & 0.91 & 0.92 & 0.95 & 0.96 & 0.90 & 0.88 & 0.90 & 0.92 & 0.99 \\
    +symbol 2 & 0.93 & 0.93 & 0.90 & 0.92 & 0.95 & 0.94 & 0.92 & 0.93 & 0.93 & 0.97 & 0.99 & 0.94 & 0.91 & 0.92 & 0.93 & 1.00 \\
    \bottomrule
    \end{tabular}
    \caption{The recall of keywords when translating IFEval English data to other languages.}
    \label{tab:recall}
\end{table*}

\section{Dataset Construction}
\label{sec:append_if_keywords}
\subsection{Rule-based Instruction Following Dataset}
We first filter out some English-specific instructions from the original dataset, such as changing the letter cases.
After filtering, the number of remaining samples is 429.
The next problem is how to extract the keywords from the translated instruction since the keywords are also translated and are required in the verification step.
We try different groups of special symbols to extract translated keywords.
The recall rates are presented in Table~\ref{tab:stat_symbol_translation}, where order 1 achieves the best performance.
Complete results across all languages are provided in Table~\ref{tab:recall}.
In addition, the number-word constraints for non-English languages are multiplied by a ratio in order to make the difficulty of the same instruction in different languages comparable.
Specifically, we calculate the ratio of the word count of English to that of a non-English language in the Flores-200 corpus using language-specific tokenizers.
we also adapt verification rules to multilingual scenarios.
For instance, word and sentence segmentation methods may vary across different languages.

During post-editing, we ask human annotators to check whether the translated keywords in the kwargs, which are used by the rule-based program, appear in the translated instruction.

% We test different groups of special symbols to extract the translated keywords.
% The recall rates are demonstrated in Table~\ref{tab:stat_symbol_translation}, where order 1 exhibits the best results.
% The complete results across all languages are shown in Table~\ref{tab:recall}.

\begin{figure*}[!t]
    \vskip 0.2in
    \centering
    \includegraphics[width=\linewidth]{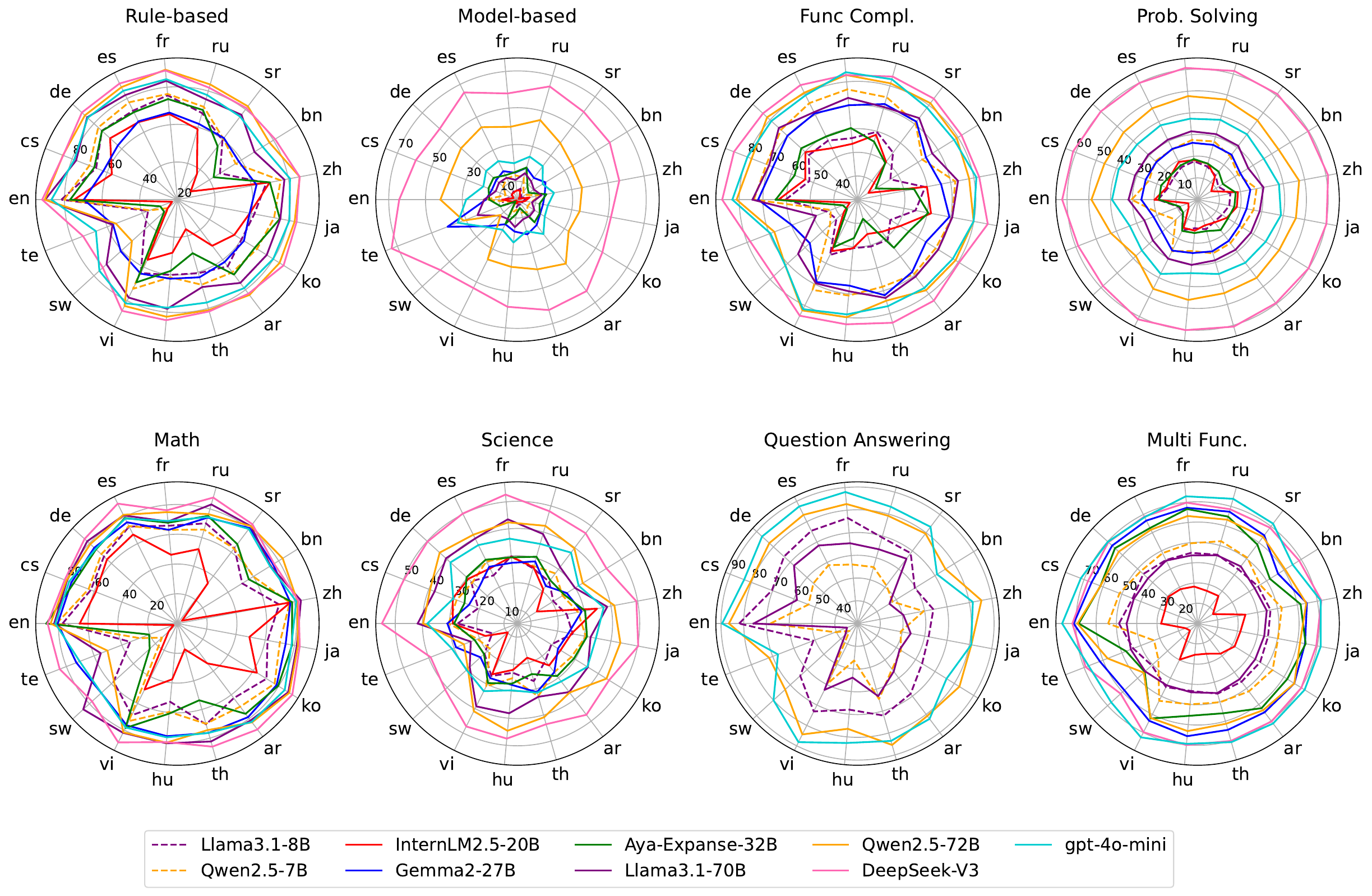}
    \caption{The radar charts visualize the performance of models on each subtask in different languages. Most model evaluated have imbalanced performance across different languages.}
    \label{fig:radar_chart}
    \vskip -0.2in
\end{figure*}

\subsection{Model-based Instruction Following Dataset}
Ten of the sixteen languages required have been provided by m-ArenaHard, which has translated the original dataset into 22 languages using Google Translate.
Based on m-ArenaHard, we further translate the English data into six other languages via Google Translate.
Subsequently, we ask human annotators to review and edit the translated instructions in all 16 languages.

\subsection{Function Completion Dataset}
The objective is to translate only the natural texts within the function comments.
However, it is challenging to prevent Google Translate from translating other elements, such as function names.
Alternatively, we instruct GPT-4o to complete this translation task with well-designed prompts~(Table~\ref{tab:translate_func_compl}).
Furthermore, a human post-editing process is employed to refine the quality of the generated translation.

\subsection{Problem Solving Dataset}
Similar to the Function Completion Dataset, we employ GPT-4o to translate the English problems into other 16 languages with a well-designed prompt~(Table~\ref{tab:translate_prob_solving}), since Google Translate cannot distinguish the parts that should remain untranslated.
Human review is also used to ensure the overall quality of the translated texts.

\subsection{Math Reasoning Dataset}
Given that the MGSM examples are written in ten languages we need, we only translate the English version into the remaining six languages via Google Translate.
This is also followed by a manual checking procedure.

\subsection{Science Reasoning Dataset}
The question and the four options of each sample are translated into 16 other languages by Google Translate.
In particular, the question and options are concatenated by option markers like ``(A)''.
After translation, we extract the translated question and options to form a new sample.
% Due to the professionalism of the testset, we require annotation personnel with a Bachelor's degree or higher to complete the annotation.

\subsection{Long-Context Question Answering Dataset}
The haystacks, needles, paragraphs and questions related to QAs are translated to other languages.
We use the parallel testsets from the UN corpus~\cite{ziemski2016united} as the haystack.
The English version contains about 128k tokens, and we extend it to other languages using Google Translate.
The sentence of the needle is also translated into 16 other languages, in which UUIDs are employed as keys and values that are not translated.
With respect to the QA data, we translate the paragraphs and questions in \textsc{XQuAD}~\cite{artetxe2020cross} to the languages we need.
Note that we also use the trick in translating IFEval to extract the answer spans.
With access to our multilingual haystacks, needles and paragraphs, we are able to synthesize the multilingual long-context testsets.

\subsection{Multiple Functions Dataset}
We only translate the user queries from English into other languages, given that the majority tool descriptions are written in English.
The user queries are initially translated by Google Translate and subsequently adjusted by human annotators.
To preserve the English parameters, we replace them with placeholders before machine translation and restore them afterward.

\section{Model Information}
\label{sec:model_info}
Here we list the evaluated models in this section.

\paragraph{Llama3.1-Instruct~\cite{dubey2024llama}} series contains three multilingual large language models with number of parameters ranging from 8B to 405B.
The pre-training corpus of Llama3.1 contains 8\% multilingual tokens, and multilingual alignment is also optimized during post-training.
In our experiments, we evaluate the 8B version and the 70B version of Llama3.1-Instruct.

\paragraph{Qwen2.5-Instruct~\cite{qwen2.5}} is a collection of multilingual language models with several sizes, ranging from 0.5B to 72B.
The models are trained with multilingual tokens in both pre-training stage and post-training stage, and are rigorously evaluated on several multilingual tasks.
In our experiments, we evaluate the 7B, 32B ,and 72B version of Qwen2.5-Instruct.

\paragraph{Aya-Expanse~\cite{dang2024aya}} is an open-weight research of models with advanced multilingual capabilities, supporting 23 languages.
The Aya Expanse 8B and 32B variants are instruction-tuned and beat Llama3.1-instruct models on the m-ArenaHard, a multilingual instruction following benchmark.

\paragraph{Gemma2-IT~\cite{team2024gemma}} family demonstrates strong multilingual capabilities, although this is not highlighted in the technical report.
We benchmark the 9B and 27B variants of Gemma2-IT.

\paragraph{InternLM2.5-chat~\cite{cai2024internlm2}} is the successor of InternLM~\cite{team2023internlm}, which is claimed as a multilingual model.
We include the 7B version and 20B version in our experiments.
InternLM2.5-7B-chat-1m is a long-context variant supporting context windows with 1M tokens.

% \paragraph{LLaMAX3-8B-Alpaca~\cite{lu2024llamax}} is a multilingual model base on Llama3-8B.
% It is continue pre-trained on corpus containing 102 languages and further fine-tuned on the English Alpaca dataset.

\paragraph{DeepSeek-V3~\cite{DeepSeekAI2024DeepSeekV3TR}} is one of state-of-the-art open-source models that achieve performance comparable to that of the best proprietary models.
It is a 671B MoE model, with 37B activated for each token.
A multilingual corpus and a multilingual-optimized tokenizer are incorporated into their training process.

\paragraph{DeepSeek-R1-Distill-Llama \& DeepSeek-R1-Distill-Qwen}~\cite{guo2025deepseek} are dense models with long reasoning capabilities, and are distilled from DeepSeek-R1 based on Llama3.1-8B, Llama3.3-70B-Instruct, Qwen2.5-Math-7B, and Qwen2.5-32B.

\paragraph{GPT-4o \& GPT-4o-mini~\cite{openai2024gpt4o}} are two of the best proprietary models that also achieve remarkable performance on multilingual tasks.
Their tokenizer can better compress multilingual texts than that of GPT-4.
GPT-4o-mini is the smaller version of GPT-4o with powerful performance.

% \paragraph{Claude-3.5-Sonnet} is also a state-of-the-art proprietary model that performs strongly on various tasks.
% It claims that Claude has the multilingual capabilities although the performonce in low-resource languages is comparatively inferior to that of high-resource ones. 

\section{Details about Prompt Templates}
\label{sec:prompts}
We present the prompt templates used in each task in this section.
Table~\ref{tab:mgsm_prompt} and Table~\ref{tab:gpqa_prompt} show the native-CoT prompts for MGSM and GPQA.
Table~\ref{tab:template} shows the prompt templates for some tasks where the original English template is used.
Table~\ref{tab:ruler_prompts} shows the prompt templates of the long-context modelling task.
Table~\ref{tab:llmjudge} shows the LLM-Judge Instruction for comparing two translations. 

\begin{table}[ht]
    \centering
    \includegraphics[width=0.5\textwidth]{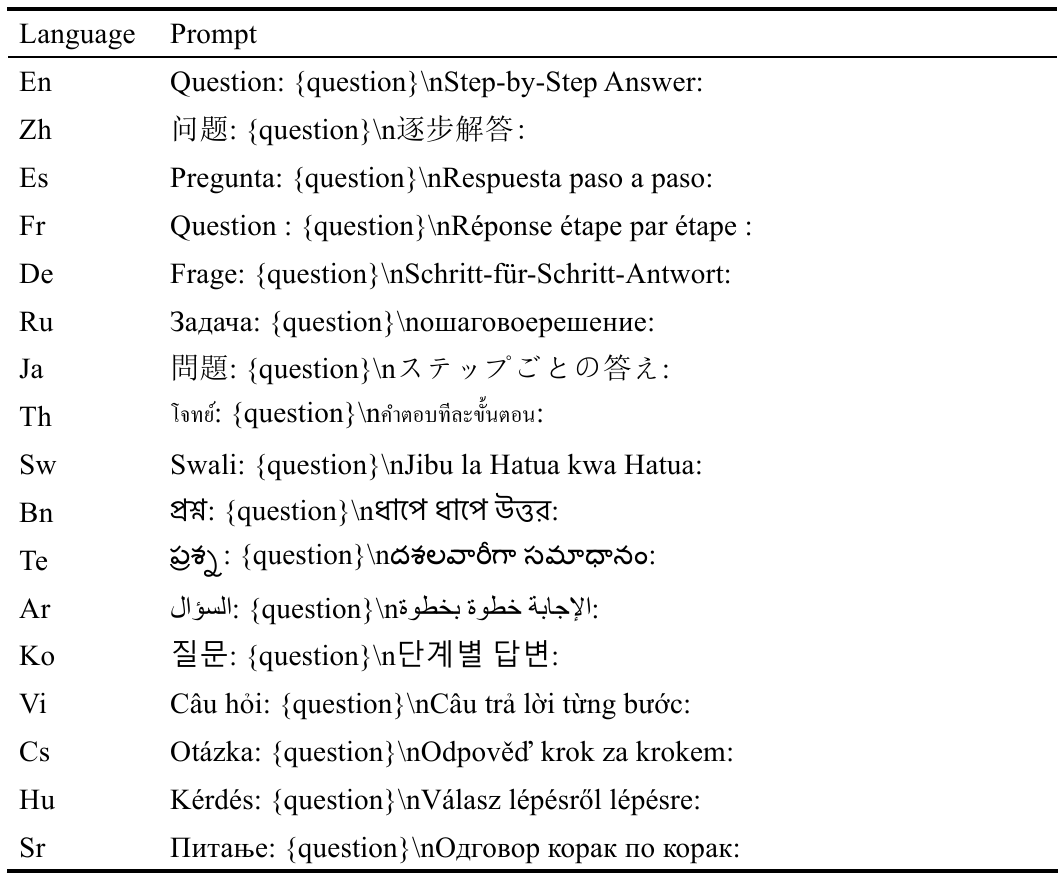}
    \caption{The native-CoT prompts of the mathematical reasoning task.}
    \label{tab:mgsm_prompt}
\end{table}

\begin{table*}[ht]
    \centering
    \includegraphics[width=0.97\textwidth]{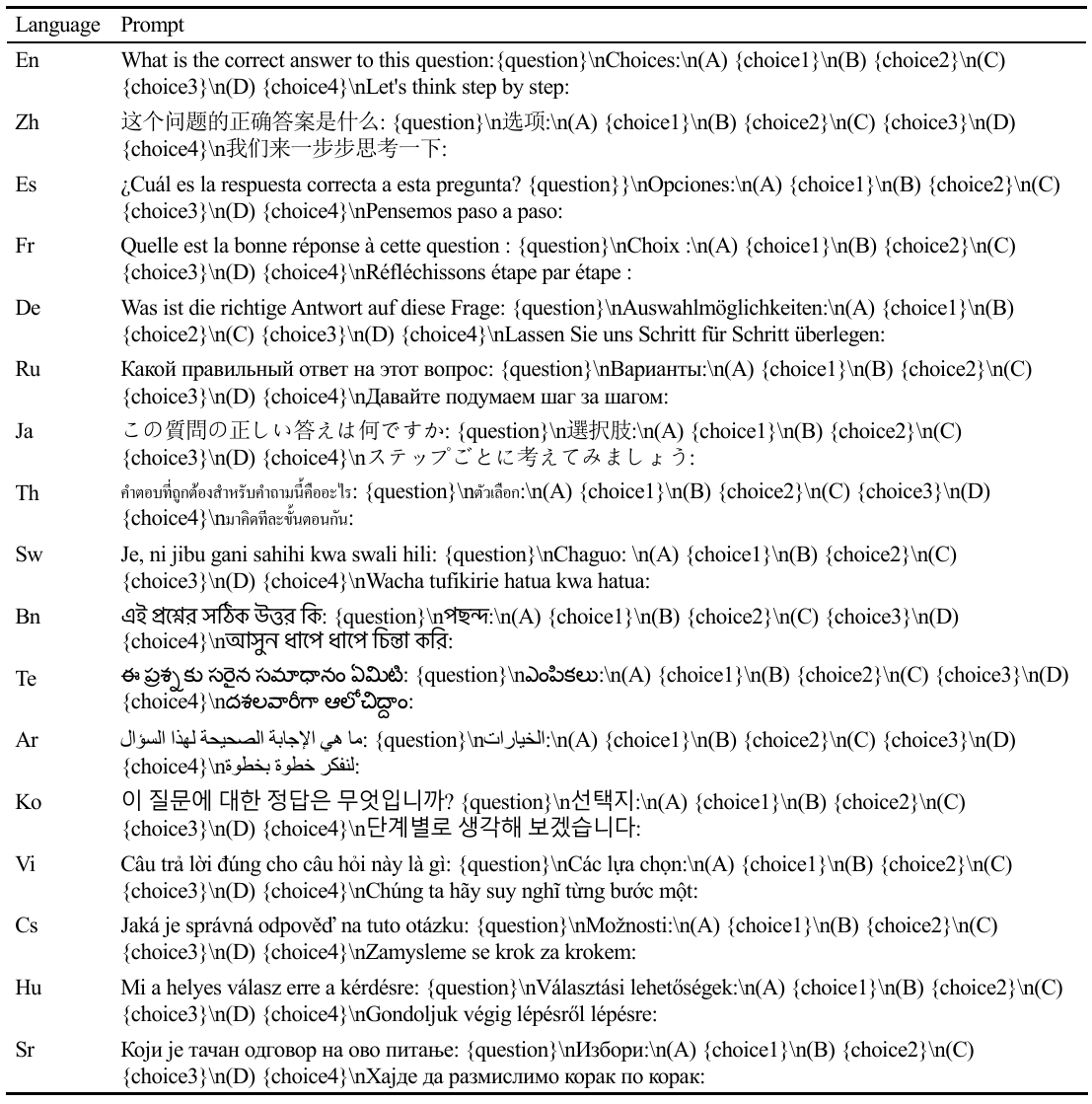}
    \caption{The native-CoT prompts of the scientific reasoning task.}
    \label{tab:gpqa_prompt}
\end{table*}

\begin{table*}[ht]
    \centering
    \small
    \begin{tabularx}{\textwidth}{l|X}
        \toprule
        Task & Prompt Template \\
        \midrule
        Rule-based instruction following & \{prompt\} \\
        \midrule
        Model-based instruction following & \{prompt\} \\
        \midrule
        Problem Solving & \textbf{\textcolor{gray}{[System Message]}}\newline
        You are an expert Python programmer. You will be given a question (problem specification) and will generate a correct Python program that matches the specification and passes all tests. You will NOT return anything except for the program.\newline
        \textbf{\textcolor{gray}{[User Message]}}\newline
        \#\#\# Question:\newline
        \{question\}\newline
        \#\#\# Format: Read the inputs from stdin solve the problem and write the answer to stdout (do not directly test on the sample inputs). Enclose your code within delimiters as follows.\newline
        \newline
        \textasciigrave\textasciigrave\textasciigrave python\newline
        \# YOUR CODE HERE\newline
        \textasciigrave\textasciigrave\textasciigrave\newline
        \newline
        \#\#\# Answer: (use the provided format with backticks)\\
        \midrule
        Function Completion & \textbf{\textcolor{gray}{[User Message]}}\newline
        Please provide a self-contained Python script that solves the following problem in a markdown code block:\newline
        \textasciigrave\textasciigrave\textasciigrave\newline
        \{prompt\}\newline
        \textasciigrave\textasciigrave\textasciigrave\newline
        \textbf{\textcolor{gray}{[Assistant Message]}}\newline
        Below is a Python script with a self-contained function that solves the problem and passes corresponding tests:\newline
        \textasciigrave\textasciigrave\textasciigrave python\\
        \midrule
        Tool use & {\textbf{\textcolor{gray}{[Tool Info]}}}\newline \{prompt\}\\
        \bottomrule         
    \end{tabularx}
    \caption{The prompt templates of the listed tasks. The prompt in the template is multilingual.}
    \label{tab:template}
\end{table*}

\begin{table*}
    \centering
    \small
    \begin{tabularx}{\textwidth}{l|X}
    \toprule
    Subtask & Prompt Template \\
    \midrule
    NIAH & \textbf{\textcolor{gray}{[User Message]}}\newline
    Some special magic uuids are hidden within the following text. Make sure to memorize it. I will quiz you about the uuids afterwards.\newline
    \{heystack\}\newline
    What are all the special magic uuids for \{query\} mentioned in the provided text?\newline
    \textbf{\textcolor{gray}{[Assistant Message]}}\newline
    The special magic uuids for \{query\} mentioned in the provided text are \\
    \midrule
    QA in a heystack (QAIAH) & \textbf{\textcolor{gray}{[User Message]}}\newline
    Answer the questions based on the given documents. Only give me the answers and do not output any other words.\newline\newline
    The following are given documents.\newline\newline
    \{context\}\newline\newline
    Answer the questions based on the given documents. Only give me the answers and do not output any other words.\newline\newline
    Questions:\newline
    \{query\}
    \textbf{\textcolor{gray}{[Assistant Message]}}\newline
    Answers:\\
    \midrule
    Variable Tracking (VT) & \textbf{\textcolor{gray}{[User Message]}}\newline
    Memorize and track the chain(s) of variable assignment hidden in the following text.\newline\newline
    \{context\}\newline
    Question: Find all variables that are assigned the value \{query\} in the text above.\newline
    \textbf{\textcolor{gray}{[Assistant Message]}}\newline
    Answer: According to the chain(s) of variable assignment in the text above, 5 variables are assgined the value \{query\}, they are: \\
    \midrule
    QA & \textbf{\textcolor{gray}{[User Message]}}\newline
    Answer the question based on the given documents. Only give me the answer and do not output any other words.\newline\newline
    The following are given documents.\newline\newline
    \{context\}\newline\newline
    Answer the question based on the given documents. Only give me the answer and do not output any other words.\newline\newline
    Question: \{query\}\newline
    \textbf{\textcolor{gray}{[Assistant Message]}}\newline
    Answer:\\
    \bottomrule
    \end{tabularx}
    \caption{The prompt templates of the long-context modelling task.}
    \label{tab:ruler_prompts}
\end{table*}

\begin{table*}[ht]
    \centering
    \begin{tabularx}{0.95\textwidth}{X}
    \toprule
    % GEMBA-SQM \\
    % \midrule
    Score the following translation from \{src\_lang\} to \{tgt\_lang\} with respect to the human reference on a continuous scale from 0 to 100 that starts with ``No meaning preserved'', goes through ``Some meaning preserved'', then ``Most meaning preserved and few grammar mistakes'', up to ``Perfect meaning and grammar''\newline
    \newline
    \{src\_lang\} source: ``\{source\}''\newline
    \{tgt\_lang\} translation: ``\{target\}''\newline
    Score:\\
    \bottomrule
    \end{tabularx}
    \caption{The GEMBA-SQM prompt.}
    \label{tab:gemba_sqm}
\end{table*}

\begin{table*}[ht]
    \centering
    \begin{tabularx}{0.95\textwidth}{X}
    \toprule
    \textbf{\textcolor{gray}{[System Message]}}\newline
    Please act as an impartial judge and evaluate the quality of the {lang} translations provided by two humans for the English source sentence displayed below. You will be given human A's translation and human B's translation. Your job is to evaluate which human's translation is better.\newline
    \newline
    You must identify and correct any mistakes or inaccurate information.\newline
    \newline
    Consider if the human's translations are accurate and fluent. Accurate means the translation conveys the same meaning, information, and nuances as the original source text. Fluent refers to the quality of the translation in terms of its naturalness, readability, and adherence to the grammatical, stylistic, and idiomatic conventions of the target language.\newline
    \newline
    Then consider whether the human's translations are consistent with the context. Code input/output and programming language syntax should not be translated. Finally, review the formatting of the translated text, including indentation, to ensure it is consistent and appropriate.\newline
    \newline
    After providing your explanation, you must output only one of the following choices as your final verdict with a label:\newline
    \newline
    1. Human A is significantly better: [[A$>>$B]]\newline
    2. Human A is slightly better: [[A$>$B]]\newline
    3. Tie, relatively the same: [[A$=$B]]\newline
    4. Human B is slightly better: [[B$>$A]]\newline
    5. Human B is significantly better: [[B$>>$A]]\newline
    \newline
    Example output: ``My final verdict is tie: [[A$=$B]]''.\newline
    \textbf{\textcolor{gray}{[User Message]}}\newline
    \textless$|$ Source Text$|$\textgreater\newline
    \{source\}\newline
    \newline
    \textless$|$The Start of Human A's Translation$|$\textgreater\newline
    \{translation\_1\}\newline
    \textless$|$The End of Human A's Translation$|$\textgreater\newline
    \newline
    \textless$|$The Start of Human B's Translation$|$\textgreater\newline
    \{translation\_2\}\newline
    \textless$|$The End of Human B's Translation$|$\textgreater\\
    \bottomrule
    \end{tabularx}
    \caption{LLM-Judge Instruction}
    \label{tab:llmjudge}
\end{table*}

\begin{table*}[ht]
    \centering
    \begin{tabularx}{0.95\textwidth}{X}
    \toprule
    \textbf{\textcolor{gray}{[System Message]}}\newline
    You are a professional translator specializing in technical content. Please translate the following English Python codes into \{tgt\_lang\}, adhering to these specific guidelines:\newline\newline
    1. **Do not translate** content representing code input/output or programming language syntax. Only translate content in comments.\newline
    2. **Maintain the original formatting** of the text, structure and indentation.\newline
    3. **Do not translate** any LaTeX code.\newline
    4. **Only output the translation** without any additional comments or explanations.\newline
    \textbf{\textcolor{gray}{[User Message]}}\newline
    \{problem\}\\
    \bottomrule
    \end{tabularx}
    \caption{Prompt for translating the Function Completion task.}
    \label{tab:translate_func_compl}
\end{table*}

\begin{table*}[ht]
    \centering
    \begin{tabularx}{0.95\textwidth}{X}
    \toprule
    \textbf{\textcolor{gray}{[System Message]}}\newline
    You are a professional translator specializing in technical content. Please translate the following English coding problems into \{tgt\_lang\}, adhering to these specific guidelines:\newline\newline
    1. **Do not translate** any LaTeX code.\newline
    2. **Do not translate** content representing code input/output or programming language syntax.\newline
    3. **Maintain the original formatting** of the text and structure.\newline
    4. **Only output the translation** without any additional comments or explanations.\newline
    \textbf{\textcolor{gray}{[User Message]}}\newline
    \{problem\}\\
    \bottomrule
    \end{tabularx}
    \caption{Prompt for translating the Problem Solving task.}
    \label{tab:translate_prob_solving}
\end{table*}

\section{Detailed results}
\label{sec:detailed_results}
Figure~\ref{fig:radar_chart} illustrates the detailed results of each model on each task.